\documentclass[lettersize,journal]{IEEEtran}

\usepackage{mathrsfs}
\usepackage{amsfonts}
\usepackage{amsmath,amssymb}
\usepackage{amsthm}
\usepackage{array}
\usepackage{url}
\usepackage{mathtools}
\usepackage{graphicx}
\usepackage{color}
\usepackage{float}
\usepackage{multirow}
\usepackage{cite}
\usepackage{CJK}
\usepackage{indentfirst}
\usepackage{subfigure}
\usepackage{ulem}
\usepackage{caption}
\usepackage{booktabs}
\usepackage{dsfont}
\usepackage{textcomp}
\usepackage{bbm}
\usepackage{pifont}
\usepackage{enumitem}
\usepackage[colorlinks,urlcolor=blue,linkcolor=blue,anchorcolor=green,citecolor=red]{hyperref}

\begin{document}
\begin{CJK}{UTF8}{gbsn}

\title{Exclusive Style Removal for Cross Domain Novel Class Discovery}
\author{Yicheng Wang, Feng Liu, Junmin Liu and Kai Sun\\
\thanks{Y. Wang is with the School of Mathematics and Statistics, Xi'an Jiaotong University, Xi'an 710049, China, and is also with the School of Mathematics and Statistics, The University of Melbourne, VIC 3010 Australia.}
\thanks{F. Liu is with the School of Computing and Information Systems, The University of Melbourne, VIC 3010 Australia.}
\thanks{J. Liu and K. Sun are with the School of Mathematics and Statistics, Xi'an Jiaotong University, Xi'an 710049, China.}
}

\markboth{Y. Wang \MakeLowercase{\textit{et al.}}: Exclusive Style Removal for Cross Domain Novel Class Discovery}%
{Y. Wang \MakeLowercase{\textit{et al.}}: Exclusive Style Removal for Cross Domain Novel Class Discovery}

\maketitle

\begin{abstract}
  As a promising field in open-world learning, \textit{Novel Class Discovery} (NCD) is usually a task to cluster unseen novel classes in an unlabeled set based on the prior knowledge of labeled data within the same domain. However, the performance of existing NCD methods could be severely compromised when novel classes are sampled from a different distribution with the labeled ones. In this paper, we explore and establish the solvability of NCD with cross domain setting under the necessary condition that the style information needs to be removed. Based on the theoretical analysis, we introduce an exclusive style removal module for extracting style information that is distinctive from the baseline features, thereby facilitating inference. Moreover, this module is easy to integrate with other NCD methods, acting as a plug-in to improve performance on novel classes with different distributions compared to the labeled set. Additionally, recognizing the non-negligible influence of different backbones and pre-training strategies on the performance of the NCD methods, we build a fair benchmark for future NCD research. Extensive experiments on three common datasets demonstrate the effectiveness of our proposed style removal strategy.
\end{abstract}


\section{Introduction}
\label{Introduction}

Generic \textit{Machine Learning} (ML), whether supervised or semi-supervised learning, typically relies on prior knowledge of a specific label space to which all samples belong. However, in open-world scenarios, it is common to encounter samples whose labels do not exist in the supervised label space of the ML model. This can significantly affect the model performance and raise concerns about model trustworthiness \cite{SSL_evaluation}. To tackle this issue, \textit{Novel Class Discovery} (NCD) \cite{NCD_survey} has been proposed and attracted significant attention in the ML community.

\par
Different from the traditional ML assumption where testing samples should all fall inside the known classes during the training process, the NCD is introduced not only to classify data into known classes but also to cluster instances that do not belong to any existing class \cite{UNO}. Specifically, given a training dataset that includes a labeled set and an unlabeled set with different label space, the goal of NCD is to learn a model that can cluster the unlabeled data by leveraging the supervised information from the labeled data, meanwhile without compromising classification performance on the labeled data \cite{ComEx}. Due to the disjoint label spaces, the labeled and unlabeled sets are often referred to as the \textit{seen} and \textit{novel} categories sets in many NCD works \cite{UNO, ComEx}. By relaxing the restriction of the same label space of semi-supervised learning, NCD becomes a promising field for open-world scenarios and various related applications such as anomaly detection \cite{SSL_evaluation}, outlier identification \cite{open_world_detection}, and so on \cite{RS, DTC}.

\par
In recent years, several NCD methods have been proposed, which can be generally categorized into one-stage and two-stage approaches \cite{NCD_survey}. Initially, NCD algorithms were usually developed by employing the two-stage strategy, which first focus on labeled data to establish a unified feature extraction framework, and then this framework is used to learn a similarity function or incorporate latent features between labeled and unlabeled data. Representative works following this approach include the \textit{Deep Transfer Clustering} (DTC) \cite{DTC}, \textit{Constrained Clustering Network} (CCN) \cite{KCL} and \textit{Meta Classification Likelihood} (MCL) \cite{MCL}.
\par
Apart from two-stage methods, the one-stage algorithms are characterized by simultaneously exploiting both labeled and unlabeled data. They typically learn a shared latent space representation with two different tasks: clustering unlabeled data and maintaining good classification accuracy on labeled sets. For example, remarkable works such as \textit{UNified Objective function} (UNO) \cite{UNO}, ComEx \cite{ComEx} and \textit{Rank Statistics} (RS) \cite{RS} all train a joint encoder with the assistance of two head modules for classification and clustering to obtain feature representations from both labeled and unlabeled data.

\par
Although many breakthroughs have been made in this field, most existing works \cite{UNO, ComEx, RS, DTC, KCL, MCL, CSDL} are introduced under the assumption that instances are consistently sampled from the same domain \cite{NCD_over_domain, SROSDA, CROW}. This assumption proves unrealistic in many real-world applications, as it inevitably results in a performance degradation for existing NCD methods when the distribution of unlabeled data differs from that of the labeled set. We construct a series of toy experiments to illustrate this issue. Note that as there is no overlap between the classes of labeled and unlabeled sets, each class obviously comes from a distinct category distribution. In contrast, the distribution mentioned in this paper refers to the low-level features of instances, or say, the data collection environments as always discussed in cross domain tasks \cite{DA_survey, DG_survey}.

First, we employ a simple corruption method (Gaussian Blur) with five increasing levels of severity to the CIFAR10 dataset \cite{CIFAR10} to create data with different distribution compared to the original dataset. Based upon the corrupted data, we then synthesize two groups of toy datasets: \textit{CIFAR10cmix} (\textbf{c}orrupt \textbf{mix}) and \textit{CIFAR10call} (\textbf{c}orrupt \textbf{all}), which respectively represent scenarios where distribution shift exists and where it is absent between labeled and unlabeled data. Further details on the settings of datasets and experiments are introduced in Sec. \ref{Motivation_Setup}.

\begin{figure}[t!]
  \centering
  \subfigure[original]{
  \includegraphics[width=0.14\linewidth]{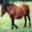}}
  \subfigure[s1]{
  \includegraphics[width=0.14\linewidth]{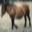}}
  \subfigure[s2]{
  \includegraphics[width=0.14\linewidth]{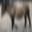}}
  \subfigure[s3]{
  \includegraphics[width=0.14\linewidth]{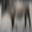}}
  \subfigure[s4]{
  \includegraphics[width=0.14\linewidth]{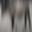}}
  \subfigure[s5]{
  \includegraphics[width=0.14\linewidth]{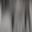}}
  \subfigure[UNO\cite{UNO}]{
  \includegraphics[width=0.3\linewidth]{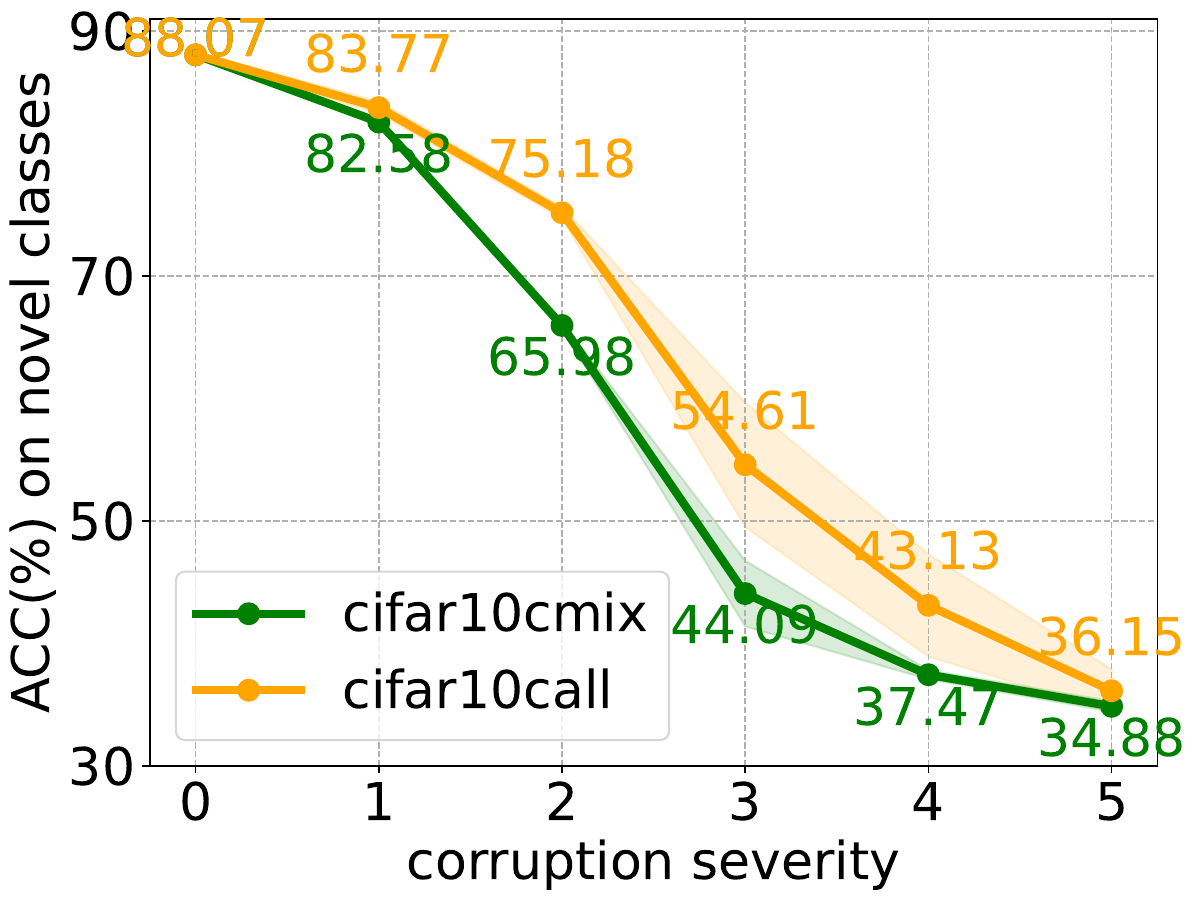}}
  \subfigure[ComEx\cite{ComEx}]{
  \includegraphics[width=0.3\linewidth]{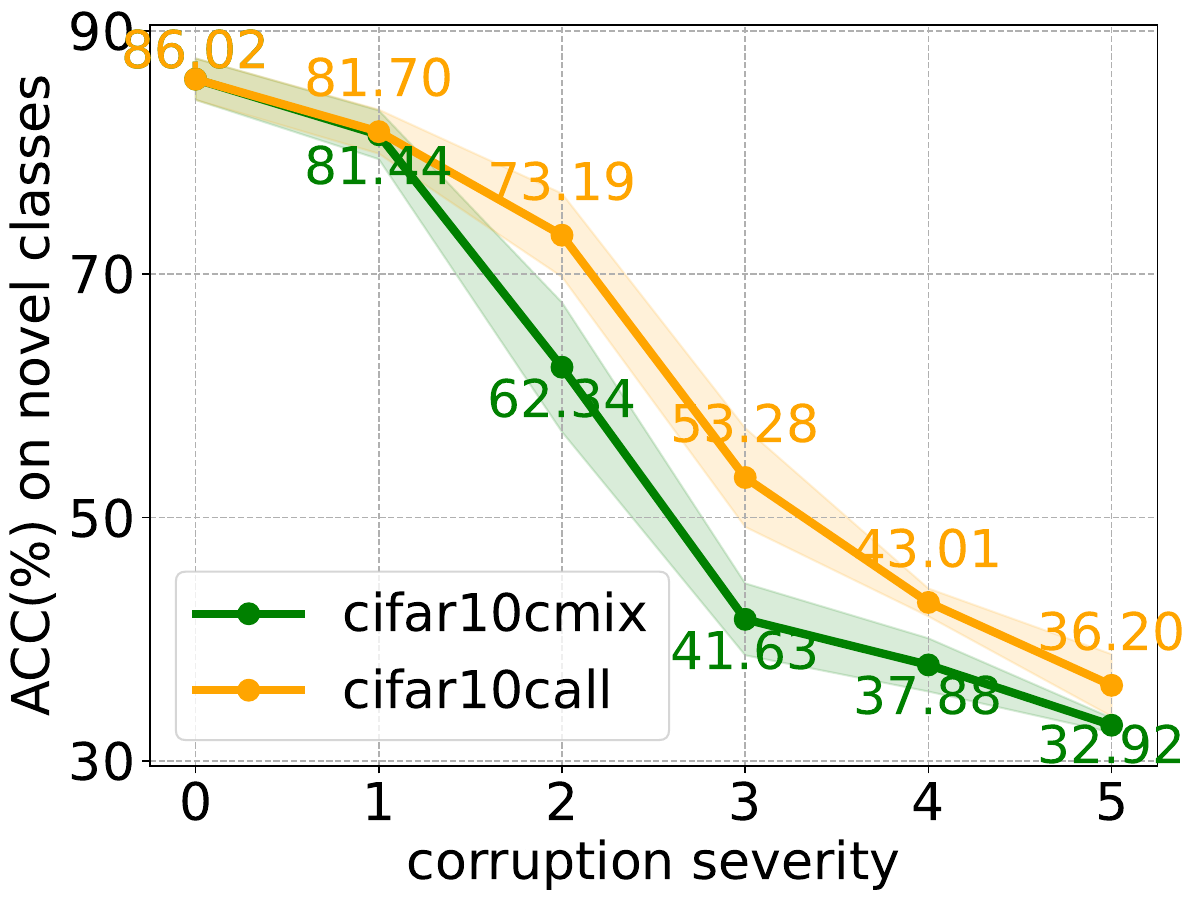}}
  \subfigure[RS\cite{RS}]{
  \includegraphics[width=0.3\linewidth]{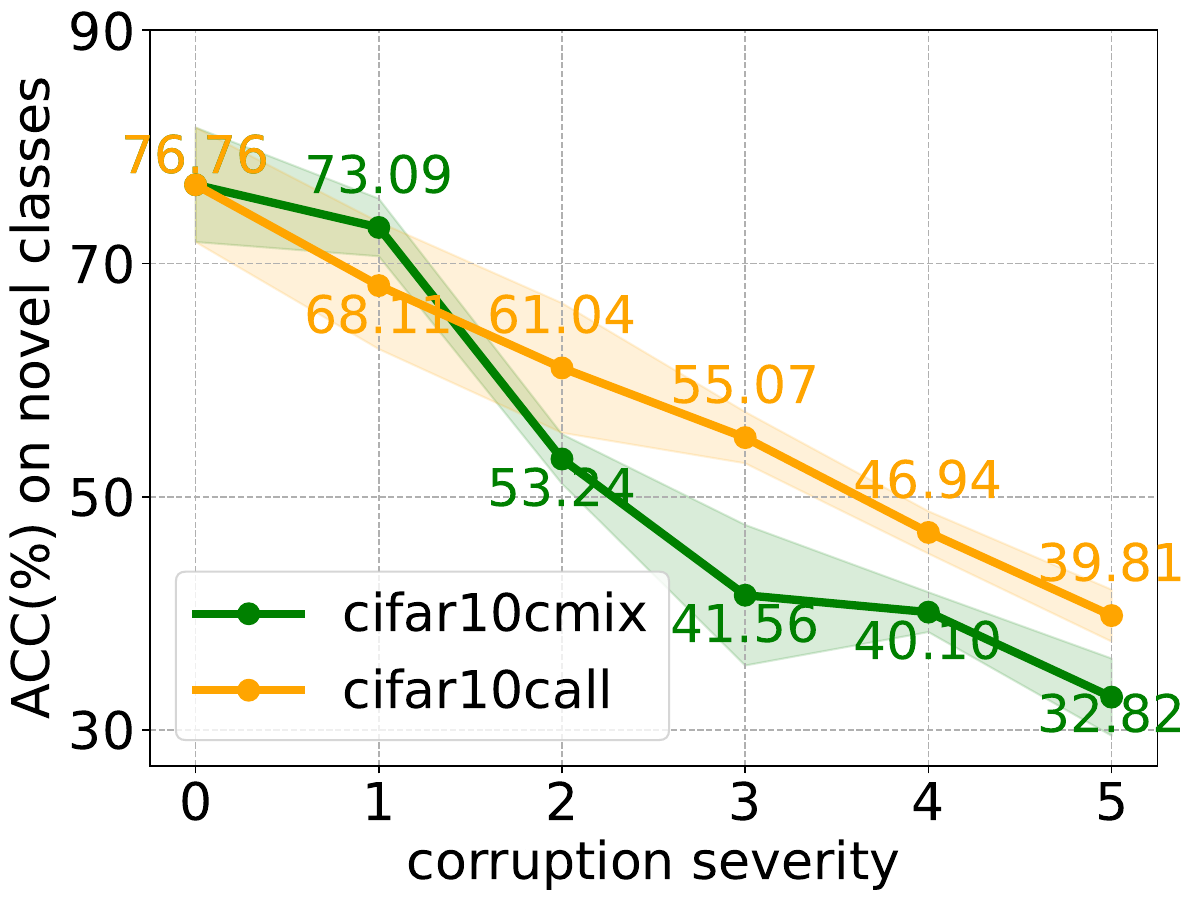}}
  \caption{(a)-(f) are one of the original instances in CIFAR10 and its corrupted versions processed by Gaussian Blur with different levels of severity (ranging from 1 to 5, termed as s1-s5), and (g)-(i) show the performance degradation of three NCD methods on \textit{novel} classes with different settings (w and w/o distribution shift).}
  \label{motivation_setup}
\end{figure}

As shown in Fig. \ref{motivation_setup}, despite the consistent performance decline of existing methods \cite{UNO, ComEx, RS} as the corruption severity increases, it is noteworthy that the performance of three methods trained on \textit{CIFAR10call} (without (w/o) distribution shift, represented by orange lines) is consistently better than that on \textit{CIFAR10cmix} (with (w) distribution shift, represented by green lines). The substantial gap observed between the two lines inspires us to address a new NCD task. Considering that the distinction between domains can be viewed as a special kind of distribution shift, and given the considerable body of works \cite{DA_survey, DG_survey} on cross domain problems, in this paper we concentrate on the task of \textit{Cross Domain NCD} (CDNCD) where \textit{unlabeled instances belong to categories and domains both different from the labeled data}.

Thus the CDNCD task is motivated by a practical perspective, as real-world ML systems need to perform well across domains and classes without any supervised information. To address this challenging task, we first expand the solvability analysis from NCD to CDNCD task and demonstrate the critical importance of removing style information for solving this new task. Based on the theory, we then introduce a solution that could built upon a series of baseline works \cite{UNO, ComEx, SimGCD} trained simultaneously with a simple yet effective style removal module. Furthermore, as the NCD field is still in its infancy, several algorithms have been proposed with diverse backbones and training settings. This results in a lack of a comprehensive and fair experimental benchmark for comparison. So we build a unified benchmark that can provide a useful reference for future NCD and related transfer learning research.

The contributions of this paper are as follows:
\begin{enumerate}
  \item We define a more challenging but practical task called \textit{Cross Domain Novel Class Discovery} by verifying the performance degradation of existing NCD methods on a series of synthesized toy datasets with distribution shift.
  \item We first analyze the solvability of CDNCD task theoretically and then propose an approach by removing the exclusive style feature between \textit{seen} and \textit{novel} data to enhance the performance of NCD methods with cross domain setting.
  \item It is found that the choice of diverse backbones and pre-training strategies have a significant impact on the performance of NCD algorithms. Therefore, a unified experimental implementary framework is developed as a fair benchmark for further research.
  \item Numerical experiments quantitatively demonstrate the effectiveness of the proposed strategy and validate its merit as a plug-in for several NCD methods.
\end{enumerate}

\section{Related Works}
\subsection{Novel Class Discovery}
The NCD problem was first introduced by Hsu \textit{et al.} from the perspective of the \textit{Transfer Learning} (TL) task and they proposed a solution termed CCN \cite{KCL}. Following this two-stage NCD work, some algorithms \cite{DTC, MCL} were developed while \cite{NCD_survey, UNO, RS} pointed out that the two-stage models only use labeled data in the first training stage which could lead to data bias.

To avoid this issue, more recent NCD works have adopted a one-stage manner to learn feature representation based on both labeled and unlabeled sets. For example, based on the same backbone \cite{ResNet} for feature extraction, UNO \cite{UNO} introduces a unified objective function for discovering \textit{novel} classes and ComEx \cite{ComEx} proposes two groups of compositional experts to enhance the discriminate capabilities to both sets.

In ML research, new directions are often defined by relaxing the assumptions or restrictions behind existing tasks to be closer to real-world applications, so is the NCD task. Generally speaking, there are two directions for expansion. One is the cross domain NCD studied here which relaxes the assumption that ``\textit{both labeled and unlabeled data come from one domain}'' \cite{NCD_over_domain}, and the other is named as \textit{Generalized Category Discovery} (GCD), which removes the limiting assumption that ``\textit{all of the unlabeled images come from new categories}'' \cite{GCD, ERL4GCD}.

As the first work to address NCD with the cross domain setting, Yu \textit{et al.} \cite{NCD_over_domain} proposed a self-labeling framework to recognize \textit{seen} classes and discover \textit{novel} categories of target domain samples simultaneously.
However, the usage of a supervised pre-trained backbone \cite{ResNet} might cause the label information leakage of \textit{novel} classes, which contradicts with vanilla setting of NCD \cite{NCD_survey}.
Meanwhile, GCD deals with NCD when the unlabeled data includes both \textit{seen} and \textit{novel} classes, without any information on the number of \textit{novel} classes \cite{PIS}. 
It is a natural extension of NCD task, requiring methods with the ability to recognize the previously \textit{seen} categories and estimate the class number of \textit{novel} classes in the unlabeled data \cite{ERL4GCD}.
Combining a representative GCD algorithm \cite{GCD} with a self-distillation mechanism and entropy regularization, SimGCD \cite{SimGCD} was introduced as an improved version and could also serve as a strong baseline in NCD tasks \cite{CSDL}.
Besides, challenging issues are recently proposed by \cite{SROSDA} and \cite{CROW} to combine the NCD and GCD tasks, which are more practical in real-world applications.

\subsection{Cross Domain Learning}
Cross domain learning consists of two well-defined tasks: \textit{Domain Adaptation} (DA) and \textit{Domain Generalization} (DG) \cite{DA_survey, DG_survey}. DA aims to transfer knowledge from a label-rich source domain to a label-scarce target domain, with the target domain data available during training \cite{DA_survey}. In contrast, the DG model is trained on multiple source domains and tested on an unseen target domain to improve the generalization ability of the model \cite{DG_survey}.

From the perspective of DA, the CDNCD defined here could be regarded as a task that relaxes the assumption that ``\textit{all data share the same category space}''. Specifically, the CDNCD model should be trained on labeled data from the source domain and unlabeled data from the target domain, which is similar to common DA methods \cite{DA_survey}, but the unlabeled data come from \textit{novel} categories that the labeled set does not belong to.

Generally, both DA and DG tasks are based on the assumption that the data from different domains share domain-invariant features suitable for discrimination \cite{partial_disentangle_DA, DA_object_detection, removing_domain_specific_DG, DASMN}. Therefore, learning to extract these domain-invariant features and removing domain-specific features is the key to solving the cross domain learning problem. 
In DA, for example, the distribution of source and target domain data is aligned by \textit{adversarial training} \cite{DASMN, Domain_adversarial, CODE}, \textit{Maximum Mean Discrepancy} (MMD) \cite{unsupervised_DA_MMD} or \textit{Optimal Transport} (OT) \cite{OT_DA} to learn the domain-invariant representation. Moreover, techniques such as reconstructing the original image to analogs in multiple domains \cite{MTAE_DG} or simply making projected textural and semantic feature orthogonal \cite{DG_HEX} encourage the DG models to focus on semantic (domain-invariant) information. 
So in this paper, we follow the idea by making the above two kinds of features with low correspondence to find and remove exclusive style features (i.e. domain-specific features) for solving the CDNCD problem.

\section{The Proposed Method}
\subsection{Motivation}
\label{Motivation_Setup}
Current NCD methods are usually based on data from a specific domain with the same distribution and may suffer when there is distribution shift between the \textit{seen} and \textit{novel} categories. To point out this issue, we first construct a series of corrupted CIFAR10 \cite{CIFAR10} with different severities of Gaussian Blur \footnote{The codes are available on: \url{https://github.com/hendrycks/robustness}}\cite{Corruptions}. Based on the original and corrupted data, two groups of toy datasets \textit{CIFAR10cmix} and \textit{CIFAR10call} are synthesized.

Specifically, in \textit{CIFAR10cmix}, the first five classes of original CIFAR10 as \textit{seen} categories are chosen as labeled data while the remaining classes of corrupted CIFAR10 are referred to as \textit{novel} categories, with their corresponding samples used as unlabeled data. In contrast, labeled and unlabeled data in \textit{CIFAR10call} are both corrupted images. Thus the \textit{CIFAR10cmix} stands for there existing distribution shift between \textit{seen} and \textit{novel} categories while the \textit{CIFAR10call} denotes the scenarios with the same distribution. Then above two groups of toy datasets are used to train and test existing NCD methods \cite{UNO, ComEx, RS}.

The results, as shown in Fig. \ref{motivation_setup}, clearly demonstrate that increasing corruption severities result in a consistent degradation in performance. More importantly, when the unlabeled data is drawn from a different distribution from that of the labeled data, as illustrated by green lines, the performance of three NCD methods on clustering \textit{novel} categories may significantly and consistently decrease compared with the same distribution setting \textit{CIFAR10call} shown by orange lines. The notable gap between two lines serves as compelling evidence that these NCD methods are sensitive to the distribution shift of data between \textit{seen} and \textit{novel} categories. This is the motivation of our work to propose the CDNCD problem and solve it to partially bridge the above gap.

\subsection{Problem Definition and Analysis of Solvability}\label{Theoretical_Analysis}

Although the definition of NCD is presented in various works \cite{NCD_survey, UNO, ComEx, RS, CSDL} in different manners, Chi \textit{et al.} \cite{Meta_Discovery} were the first to provide the formal definition and theoretical solvability of the NCD problem. They clarify the assumptions underlying NCD that high-level semantic features should be shared between labeled and unlabeled data.

Building on the concepts introduced in \cite{Meta_Discovery}, we define and outline the assumptions of CDNCD problem and discuss its solvability. Our analysis leads to the conclusion that in addition to the requirement for similar semantic information between labeled and unlabeled set, solving the CDNCD also hinges on the removal of the exclusive style information induced by the cross domain setting.

Two crucial definitions from \cite{Meta_Discovery} regarding the $K$-$\epsilon$-separable \textit{random variable} (r.v.) and the consistent $K$-$\epsilon$-separable transformation set are list as follows:

\textbf{Definition 1} ($K$-$\epsilon$-separable r.v.)\textbf{.}
\textit{
  Given a r.v. $X \sim \mathbb{P}_{X}$ defined on space $\mathcal{X}\subset \mathbb{R}^d$, $X$ is $K$-$\epsilon$-separable with a non-empty function set $\mathcal{F}=\{f: \mathcal{X} \rightarrow \mathcal{I}\}$, if $\forall f \in \mathcal{F}$,
  \begin{equation}
    \begin{aligned}
      &\tau(X, f(X))\\
      &:=\max _{i, j \in \mathcal{I}, i \neq j} \mathbb{P}_{X}\left(R_{X \mid f(X)=i} \cap R_{X \mid f(X)=j}\right)=\epsilon,
    \end{aligned}
  \end{equation}
  where $\mathcal{I} = \{i_1, \cdots, i_K\}$ is an index set, $f(X)$ is an induced r.v. whose source of randomness is $X$, and $R_{X \mid f(X)=i}$ is the support set of $\mathbb{P}_{X \mid f(X)=i}$.
}

\textbf{Definition 2} (Consistent $K$-$\epsilon$-separable Transformation Set)\textbf{.}
\textit{
  Given the r.v. $X \sim \mathbb{P}_{X}$ that is $K$-$\epsilon$-separable with $\mathcal{F}$, a transformed r.v. $\pi(X)$ is $K$-$\epsilon$-separable with $\mathcal{F}$, if $\forall f \in \mathcal{F}$,
  \begin{equation}
    \begin{aligned}
      &\tau(\pi(X), f(X))\\
      &:=\max _{i, j \in \mathcal{I}, i \neq j} \mathbb{P}_{\pi(X)}\left(R_{\pi(X) \mid f(X)=i} \cap R_{\pi(X) \mid f(X)=j}\right)=\epsilon,
    \end{aligned}
  \end{equation}
  where $\pi: \mathcal{X} \rightarrow \mathbb{R}^{d_{r}}$ ($d_{r}\ll d$) is a dimension reduction transformation function. Then, a non-empty set $\Pi$ is a consistent $K$-$\epsilon$-separable transformation set satisfying that $\forall \pi \in \Pi$, $\pi(X)$ is $K$-$\epsilon$-separable with $\mathcal{F}$.
}

Similar to the assumptions in cross domain learning works \cite{partial_disentangle_DA, removing_domain_specific_DG}, here we also assume that image data follow a joint distribution of style and content information, denoted as $X \sim \mathbb{P}_{X} = \mathbb{P}_{sX} \mathbb{P}_{cX}$, where $\mathbb{P}_{s}$ and $\mathbb{P}_{c}$ stand for the margin distributions of style and content, respectively. Consequently, the features processed by a non-linear transformation $\pi \in \Pi$ can theoretically be decomposed into two parts: $\pi(X) = [\pi_c(X),\pi_s(X)]$, where $[\cdot , \cdot]$ is tensors concatenation, and $\pi_c(X), \pi_s(X) \in \mathbb{R}^{d_{r}}$ represent the corresponding content feature and style feature. Note that it is not necessary for these two features to have the same dimension, while they are set equally to $d_{r}$ here just for simplicity in the following analysis and data processing.

Building on \cite{Meta_Discovery}, the NCD problem on cross domain setting can be defined as follows. The difference is that the transformation set $\Pi$ is replaced by $\Pi_c$ for dimension reduction of the content feature.

\textbf{Definition 3} (CDNCD)\textbf{.}
\textit{In data-label joint distribution $\{\mathcal{X},\mathcal{Y}\}$, two r.v. $X^{l}$, $X^{u}$ are sampled from $\mathcal{X}^l$ and $\mathcal{X}^u$ to represent labeled and unlabeled data respectively, where $X^{l} \sim \mathbb{P}_{X^{l}} = \mathbb{P}_{cX^{l}} \mathbb{P}_{sX^{l}}$ and $X^{u} \sim \mathbb{P}_{X^{u}} = \mathbb{P}_{cX^{u}} \mathbb{P}_{sX^{u}}$, the classification function for data with ground-truth labels $f^l: \mathcal{X} \rightarrow \mathcal{Y}^l$ and a function set $\mathcal{F}=\{f: \mathcal{X} \rightarrow \mathcal{Y}^u\}$, where $\mathcal{Y}^l = \{i^{l}_1, \ldots, i^{l}_{K^{l}}\}$ and $\mathcal{Y}^u = \{i^{u}_1, \ldots, i^{u}_{K^{u}}\}$. Then we have the following assumptions:
\begin{enumerate}[label=\textbf{(\Alph*)},leftmargin=*]
  \item The support set of $X^{l}$ and the support set of $X^{u}$ are disjoint, and underlying classes of $X^{l}$ are different from those of $X^{u}$ (i.e., $\mathcal{Y}^l \bigcap \mathcal{Y}^u = \emptyset$), and $\mathbb{P}_{sX^{l}}\neq \mathbb{P}_{sX^{u}}$;\label{assumption_a}
  \item $X^{l}$ is $K^l$-$\epsilon^l$-separable with $\mathcal{F}^l=\{f^l\}$ and $X^{u}$ is $K^u$-$\epsilon^u$-separable with $\mathcal{F}^u$, where $\epsilon^l = \tau(X^l, f^l(X^l))<1$ and $\epsilon^u =\min_{f\in\mathcal{F}}\tau(X^u, f(X^u))<1$;\label{assumption_b}
  \item There exist a consistent $K^l$-$\epsilon^l$-separable transformation set $\Pi^l_c$ for $X^{l}$ and a consistent $K^u$-$\epsilon^u$-separable transformation set $\Pi^u_c$ for $X^{u}$;\label{assumption_c}
  \item $\Pi^l_c \bigcap \Pi^u_c \neq \emptyset$.\label{assumption_d}
\end{enumerate}
With above assumptions \ref{assumption_a}-\ref{assumption_d} hold, the goal of CDNCD is to learn a dimension reduction transformation $\hat{\pi}_c: \mathcal{X} \rightarrow \mathbb{R}^{d_{r}}$ via minimizing $\mathcal{J}(\hat{\pi}_c)=\tau\left(\hat{\pi}_c\left(X^{l}\right), f^{l}\left(X^{l}\right)\right)+\tau\left(\hat{\pi}_c\left(X^{u}\right), f^{u}\left(X^{u}\right)\right)$ such that $\hat{\pi}_c\left(X^{u}\right)$ is $K^u$-$\epsilon^u$-separable, where $f^{u} \in \mathcal{F}$ and $d_r\ll d$.
}

The interpretation for \ref{assumption_a}-\ref{assumption_d} are the same as those in \cite{Meta_Discovery}, with the addition of a supplement to \ref{assumption_a}: $\mathbb{P}_{sX^{l}}\neq \mathbb{P}_{sX^{u}}$ implies that the style distribution of $X^{l}$ and $X^{u}$ is different. In other words, the labeled and unlabeled data come from different domains.

\textbf{Theorem 1} (CDNCD is Theoretically Solvable)\textbf{.} 
\textit{
  Given $X^{l}$, $X^{u}$, $f^l$ and $\mathcal{F}$ defined above and assumptions \ref{assumption_a}-\ref{assumption_d} hold, then $\hat{\pi}_c$ is $K^u$-$\epsilon^u$-separable. If $\epsilon^u = 0$, then CDNCD is theoretically solvable.
}

Theorem 1 suggests that on the CDNCD setting, it is possible to learn a suitable transformation $\hat{\pi}_c$ to achieve separable content features for inference. The proof of Theorem 1 is similar as that in \cite{Meta_Discovery}, so it is omitted here. The only difference lies in replacing the transformation set $\Pi$ with $\Pi_c$. In addition, a theorem regarding that CDNCD is not solvable when condition \ref{assumption_d} does not hold is similar to that in \cite{Meta_Discovery}. The latter argues that the consistent semantic information between labeled and unlabeled data is a necessary condition for solving the NCD problem, so is the CDNCD.

When we totally following the way of NCD setting \cite{Meta_Discovery}, ignoring to remove the exclusive style information caused by the cross domain context, the CDNCD problem might be unsolvable. This claim is supported by a Impossibility Theorem presented formally below.

\textbf{Theorem 2} (Impossibility Theorem with Style Information)\textbf{.} 
\textit{
  Given solvable CDNCD problem with $K^u$-$\epsilon^u$-separable transformation set $\hat{\pi}_c$. $X^{l}$, $X^{u}$, $f^l$ and $\mathcal{F}$ defined above and assumptions \ref{assumption_a}-\ref{assumption_d} hold. Consider conditions below on the expanded transformation set $\Pi = [\Pi_c, \Pi_s]:= \{[\pi_c, \pi_s], \pi_c \in \Pi_c  \ {\rm and} \  \pi_s \in \Pi_s \}$ as follows:
  \textbf{(C*)} There exist a consistent $K^l$-$\epsilon^l$-separable transformation set $\Pi^l$ for $X^{l}$ and a consistent $K^u$-$\epsilon^u$-separable transformation set $\Pi^u$ for $X^{u}$;
  \textbf{(D*)} $\Pi^l \bigcap \Pi^u \neq \emptyset$. 
  By utilizing conditions \ref{assumption_a}-\ref{assumption_d} in \textbf{Definition 3}, \textbf{(C*)} can be hold, while \textbf{(D*)} might not be achievable. This implies that $\hat{\pi}\in \Pi$ might not be $K^u$-$\epsilon^u$-separable.
}

Theorem 2 demonstrates that solving the CDNCD might be impossible without removing the style feature $\hat{\pi}_s(X)$. In other words, if the goal is to find a transformation $\hat{\pi}: \mathcal{X} \rightarrow \mathbb{R}^{2dr}$, where $\hat{\pi}(X) = [\hat{\pi}_c(X),\hat{\pi}_s(X)]$ includes both content and style features for the dimension reduction of data, then the CDNCD might be ill-defined.

The proof of Theorem 2 is partially based on a lemma listed below.

\textbf{Lemma} (Dimension Lemma of $K$-$\epsilon$-separable r.v.)\textbf{.}
\textit{
  Given a $d$-dimension bounded space $\mathcal{X} \subset \mathbb{R}^d$, an index set $\mathcal{I} = \{i_1,...,i_K\}$, a $n$-dimension subspace $\mathcal{W}\subset \mathbb{R}^n$ and a $m$-dimension subspace $\mathcal{Z}\subset \mathbb{R}^m$, and $\mathcal{Z} \subset \mathcal{W} \subset \mathcal{X}$ with $m<n<d$, then $K$-$\epsilon$-separable r.v. $Z \in \mathcal{Z}$ with $\mathcal{F}=\{f: \mathcal{X} \rightarrow \mathcal{I}\}$ is a sufficient but not necessary condition for $K$-$\epsilon$-separable r.v. $W \in \mathcal{W}$ with the same $\mathcal{F}$.
}

The proof of this Lemma is available in the Appendix \ref{proof_lemma}.

Based on this lemma, it is evident that the original assumption \ref{assumption_c} is a sufficient but not necessary condition for the assumption \textbf{(C*)} to hold. We only need to prove that the condition $\Pi^l_c \bigcap \Pi^u_c \neq \emptyset$ could not guarantee $\Pi^l \bigcap \Pi^u \neq \emptyset$. The proof of this assert is provided in Appendix \ref{proof_Theorem_2}.

\subsection{Model Overview}

Based on the analysis in \ref{Theoretical_Analysis}, our goal is to ensure that the content feature $\pi_c(X)$ remains uncorrelated with the style feature $\pi_s(X)$, as the latter does not contribute to class prediction. Therefore, $\pi_s(X)$ need to be removed from the transformed feature $\pi(X)$ to theoretically guarantee the solvability of the CDNCD. In practice, this decoupling of the non-linear transformation $\pi \in \Pi$ into two independent parts can be achieved using two parallel deep neural networks and rational regularization \cite{DA_object_detection}. This approach enables the alignment of the feature distribution $\pi_c(X)$ between labeled and unlabeled sets for inference.

\begin{figure}[htbp]
  \centering
  \includegraphics[width=1\linewidth]{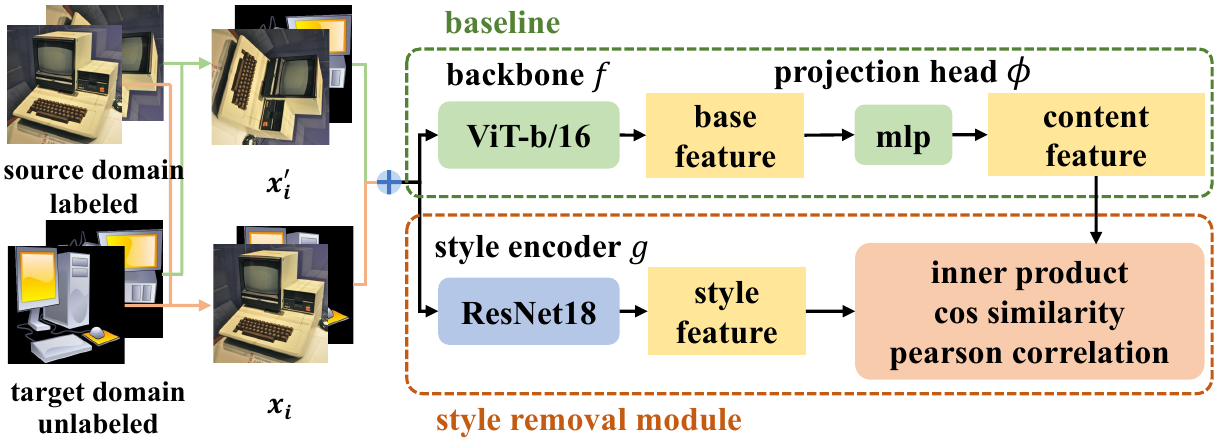}
  \caption{The framework of the proposed method. The core elements in framework of diverse baselines such as backbone and projection head, are shown in the green dashed box, while the details of baselines during training and inference are omitted for simplicity.}
  \label{model_overview}
\end{figure}

So our proposed method consists of two components: a baseline NCD work (such as \cite{UNO}, \cite{ComEx} and \cite{SimGCD}) and an exclusive style removal module called the style encoder. As shown in Fig. \ref{model_overview}, these two parallel models are trained simultaneously to separate content and style features. During inference, the base feature is fed to the classification head as same as the original baseline to predict the output labels directly.

\subsubsection*{Baseline Model}
Here we select three representative NCD approaches termed UNO \cite{UNO}, ComEx \cite{ComEx} and SimGCD \cite{SimGCD} as baseline models. Specifically, based on \cite{SimGCD}, we employ a vision transformer ViT-b \cite{ViT} pre-trained on ImageNet in a self-supervised manner \cite{DINO} as a unified feature representation backbone termed $f$. Unlike \cite{NCD_over_domain, SROSDA, CROW}, this self-supervised strategy does not contradict with the NCD setting, as the \textit{novel} classes have no supervised information throughout all training processes. The structure and pre-training procedures of the backbone $f$ in baselines \cite{UNO, ComEx, SimGCD} all follow the same manner as \cite{SimGCD}. Besides, the projection and classification head are consistent with corresponding baseline methods \cite{UNO, ComEx, SimGCD} respectively. The reason why an unified backbone is chosen will be discussed in next section.

For instance, here we adopt SimGCD \cite{SimGCD} as a baseline. Specifically, unsupervised contrastive loss and supervised contrastive loss are used to fine-tune the backbone $f$ and projection head $\phi$. Formally, two views $\boldsymbol{x}_{i}$ and $\boldsymbol{x}_{i}^{\prime}$ with different random augmentation are processed by the $f$ and $\phi$ to generate two feature representations $\boldsymbol{z}_{i} = \phi\left(f\left(\boldsymbol{x}_{i}\right)\right)$ and $\boldsymbol{z}_{i}^{\prime} = \phi\left(f\left(\boldsymbol{x}_{i}^{\prime}\right)\right)$. The unsupervised contrastive loss for representation learning is defined as:
\begin{equation}
  \mathcal{L}_{\text {rep}}^{u}=\frac{1}{|B|} \sum_{i \in B}-\log \frac{\exp \left(\boldsymbol{z}_{i}^{\top} \boldsymbol{z}_{i}^{\prime} / \tau_{u}\right)}{\sum_{n}^{n \neq i} \exp \left(\boldsymbol{z}_{i}^{\top} \boldsymbol{z}_{n}^{\prime} / \tau_{u}\right)},
\end{equation}
in which $\tau_{u}$ is a temperature hyper-parameter for unsupervised contrastive loss, $B$ is the batch of data including labeled and unlabeled samples.

To effectively leverage existing label information, the outside version of supervised contrastive loss \cite{SupCon} is added as follows.
\begin{equation}
\mathcal{L}_{\text {rep}}^{s}=\frac{1}{|B^{l}|} \sum_{i \in B^{l}}\frac{1}{|\mathcal{N}(i)|} \sum_{q \in \mathcal{N}(i)} -\log \frac{\exp \left(\boldsymbol{z}_{i}^{\top} \boldsymbol{z}_{q}^{\prime} / \tau_{c}\right)}{\sum_{n}^{n \neq i} \exp \left(\boldsymbol{z}_{i}^{\top} \boldsymbol{z}_{n}^{\prime} / \tau_{c}\right)},
\end{equation}
where $\mathcal{N}(i)$ is the set of negative samples that hold the same label as the $i$-th sample in the labeled batch $B^{l}$, $\tau_{c}$ is a temperature hyper-parameter for supervised contrastive loss. Thus representation loss is defined as $\mathcal{L}_{\text{rep}} = \left(1 - \lambda\right)\mathcal{L}_{\text {rep}}^{u} + \lambda\mathcal{L}_{\text {rep}}^{s}$, where $\lambda$ is set to balance the two losses.

Instead of the self-labeling strategy employed in \cite{UNO, ComEx}, SimGCD \cite{SimGCD} used self-distillation as a parametric classification paradigm which consists of student and teacher networks to further enhance the representative capability of the model.
Based on the latent feature $\boldsymbol{h}_{i} = f\left(\boldsymbol{x}_{i}\right)$ and randomly initialized prototypes $\mathcal{C}=\left\{\boldsymbol{c}_{1}, \ldots, \boldsymbol{c}_{K}\right\}$, where $K = |\mathcal{Y}^l\bigcup \mathcal{Y}^u|$ is the total number of categories. Then the soft label for each augmented sample is $\boldsymbol{p}_{i} = (\boldsymbol{p}_{i}^{(1)},...,\boldsymbol{p}_{i}^{(K)})^{\top}$ in which every element $\boldsymbol{p}_{i}^{(k)}$ is computed by:
\begin{equation}
  \boldsymbol{p}_{i}^{(k)}=\frac{\exp \left(\frac{1}{\tau_{s}}\left(\boldsymbol{h}_{i} /\left\|\boldsymbol{h}_{i}\right\|_{2}\right)^{\top}\left(\boldsymbol{c}_{k} /\left\|\boldsymbol{c}_{k}\right\|_{2}\right)\right)}{\sum_{k^{\prime}} \exp \left(\frac{1}{\tau_{s}}\left(\boldsymbol{h}_{i} /\left\|\boldsymbol{h}_{i}\right\|_{2}\right)^{\top}\left(\boldsymbol{c}_{k^{\prime}} /\left\|\boldsymbol{c}_{k^{\prime}}\right\|_{2}\right)\right)},
\end{equation}
where $\tau_{s}$ is a temperature for student network. For another view $\boldsymbol{x}_{i}^{\prime}$, the soft label $\boldsymbol{q}_{i}^{\prime}$ is computed by teacher network with $\tau_{t}$ similarly. So the unsupervised cluster objective with mean-entropy maximum regularization term \cite{MSN_selfdistillation} is defined as:

\begin{equation}
  \mathcal{L}_{\mathrm{cls}}^{u}=\frac{1}{|B|} \sum_{i \in B} \ell\left(\boldsymbol{q}_{i}^{\prime}, \boldsymbol{p}_{i}\right)-\varepsilon H(\overline{\boldsymbol{p}}),
\end{equation}
where $\ell$ is the \textit{Cross Entropy} (CE) loss function, $H\left(\cdot\right)$ is the entropy function and $\overline{\boldsymbol{p}}=\frac{1}{2|B|} \sum_{i \in B}\left(\boldsymbol{p}_{i}+\boldsymbol{p}_{i}^{\prime}\right)$ indicates the average prediction of a mini-batch.

In order to guarantee the performance on labeled data, the general CE loss is used as a supervised objective defined as:

\begin{equation}
  \mathcal{L}_{\mathrm{cls}}^{s}=\frac{1}{\left|B^{l}\right|} \sum_{i \in B^{l}} \ell\left(\boldsymbol{y}_{i}, \boldsymbol{p}_{i}\right),
\end{equation}
where $\boldsymbol{y}_{i}$, $\boldsymbol{p}_{i}$ are the ground truth and predicted label of the $i$-th sample, and $B^{l}$ is the batch of labeled data. Then the classification loss is set as $\mathcal{L}_{\mathrm{cls}} = \left(1 - \lambda\right)\mathcal{L}_{\mathrm{cls}}^{u} + \lambda\mathcal{L}_{\mathrm{cls}}^{s}$.

\subsubsection*{Exclusive Style Removal Module}
To ensure the solvability of NCD on cross domain setting as discussed in \ref{Theoretical_Analysis}, we propose a simple yet effective strategy for better aligning content feature. This strategy involves using a ResNet18 \cite{ResNet} trained with baseline simultaneously for extracting feature of style information that is distinctive from the discriminative feature obtained from the backbone and projection head for classification.

To validate this statement, we separately use three common similarity measures as objective functions to assess the correspondence between content feature for inference $\boldsymbol{z}_{i} =  \phi\left(f\left(\boldsymbol{x}_{i}\right)\right)$ and style feature extracted by the style encoder $\boldsymbol{v}_{i} = g\left(\boldsymbol{x}_{i}\right)$. These measures include the inner product $\boldsymbol{z}_{i}^{\top} \boldsymbol{v}_{i}$, cosine similarity $\frac{\boldsymbol{z}_{i}^{\top} \boldsymbol{v}_{i}}{\|\boldsymbol{z}_{i}\|_{2} \|\boldsymbol{v}_{i}\|_{2}}$ and Pearson correlation $\frac{\text{cov}(\boldsymbol{z}_{i}, \boldsymbol{v}_{i})}{\sigma_{\boldsymbol{z}_{i}} \sigma_{\boldsymbol{v}_{i}}}$, where $\text{cov}\left(\cdot,\cdot\right)$ is the covariance and $\sigma_{\boldsymbol{z}_{i}}$ is the standard deviation of $\boldsymbol{z}_{i}$. To ensure that the style and content feature are distinct from each other, three different style removal objective functions minimized during training are defined respectively:

\begin{equation}
  \mathcal{L}_{\mathrm{orth}} = \text{abs}\left(\boldsymbol{z}_{i}^{\top} \boldsymbol{v}_{i}\right),
  \label{L_orth}
\end{equation}
\begin{equation}
  \mathcal{L}_{\mathrm{cossimi}} = \text{abs}\left(\frac{\boldsymbol{z}_{i}^{\top} \boldsymbol{v}_{i}}{\|\boldsymbol{z}_{i}\|_{2} \|\boldsymbol{v}_{i}\|_{2}}\right),
  \label{L_cossimi}
\end{equation}
\begin{equation}
  \mathcal{L}_{\mathrm{corr}} = \text{abs}\left(\frac{\text{cov}(\boldsymbol{z}_{i}, \boldsymbol{v}_{i})}{\sigma_{\boldsymbol{z}_{i}} \sigma_{\boldsymbol{v}_{i}}}\right),
  \label{L_corr}
\end{equation}
where $\text{abs}(\cdot)$ is the absolute value function.

We use a unified format to define the style removal function as follows:
\begin{equation}
  \mathcal{L}_{\mathrm{style\_removal}} = \lambda_a \mathcal{L}_{\mathrm{orth}} + \lambda_b\mathcal{L}_{\mathrm{cossimi}} + \lambda_c\mathcal{L}_{\mathrm{corr}},
\end{equation}
where value of $\lambda_a$, $\lambda_b$ and $\lambda_c$ are set to 0 or 1 to control the usage of different functions such that $\lambda_a + \lambda_b + \lambda_c = 1$.
Following the baseline \cite{SimGCD}, the overall loss function is defined as $\mathcal{L} = \mathcal{L}_{\text{rep}} + \mathcal{L}_{\mathrm{cls}} + w\mathcal{L}_{\mathrm{style\_removal}}$, where $w$ is a hyper-parameter to balance the style removal loss and baseline loss.

Simialrly, as for baseline UNO \cite{UNO} and ComEx \cite{ComEx}, according to Fig. \ref{model_overview}, the \textit{style encoder} $g$ is integrated to the backbones $f$ with projection head $\phi$ of these NCD methods as a parallel plug-in module and the style removal function $\mathcal{L}_{\mathrm{style\_removal}}$ is added to the original baseline loss in the same way.

\subsection{Lack of Fairly Comparable Benchmark}
Given that the field of NCD is still in its early stages, there is currently no standard benchmark for fair comparison, as different works have used diverse backbones and pre-trained strategies \cite{NCD_survey, clevr4}. For example, RS \cite{RS}, UNO \cite{UNO}, and ComEx \cite{ComEx} employed vanilla ResNet18 \cite{ResNet} to learn a unified feature extractor based on the training data in a specific task in a self-supervised manner. Besides, DualRS \cite{Dual_RS} utilized ResNet50 \cite{ResNet} pre-trained on ImageNet via self-supervision MoCov2 \cite{MoCov2}. In contrast, GCD \cite{GCD} and SimGCD \cite{SimGCD} used self-supervised DINO \cite{DINO} to pre-train the ViT \cite{ViT} backbone on ImageNet for feature extractor. Even in \cite{NCD_over_domain, SROSDA} and \cite{CROW}, supervised pre-trained ResNet50 \cite{ResNet} and CLIP \cite{CLIP} were respectively used as feature extractor, which could potentially lead to label information leakage for \textit{novel} class samples. These models actually violates the NCD setting, as many \textit{novel} categories in the training dataset might have already been seen in large scale datasets with labels or prompt texts.

In the downstream applications of deep learning research, it is well known that performance is highly dependent on the backbone networks and corresponding pre-trained strategy for base feature extraction \cite{ResNet, ViT}. Different backbones and pre-trained manners might lead to significantly different results and data bias \cite{clevr4}, regardless of the outcome achieved by modules hand-crafted specially for specific tasks. Therefore, we designed a series of warm-up experiments to compare the performance of two NCD methods \cite{UNO, ComEx} with different pre-trained backbones. The detailed results and analysis are presented in Sec. \ref{Benchmark_Setup}.

\section{Experiments}
\subsection{Experimental Setup}
\textbf{Datasets.} In our experiments, we use three datasets: CIFAR10 \cite{CIFAR10}, OfficeHome \cite{OfficeHome} and DomainNet40 \cite{DomainNet, DomainNet40}. As for CIFAR10,  following the same setting of existing NCD tasks, we utilize the first five classes as labeled data and the remaining five classes as unlabeled sets.

The original OfficeHome is a image dataset designed for DA and DG tasks in computer vision \cite{DA_survey, DG_survey}. It consists of images from four different domains: \textit{Art} (\textbf{A}), \textit{Clipart} (\textbf{C}), \textit{Product} (\textbf{P}), and \textit{Real-World} (\textbf{R}). Each domain contains 65 classes of images with various office-related objects and scenes. We use the first 40 classes for experiments and split 20:20 for labeled and unlabeled data. By sequentially combining each pair of domains from four domains as labeled and unlabeled datasets, we establish twelve experiment settings for the cross domain conditions. Following the naming conventions in cross domain tasks \cite{DA_survey, DG_survey}, the term $\textbf{R}\rightarrow\textbf{A}$ indicates the labeled data comes from the \textit{Real-World} domain and the unlabeled data are sampled from the \textit{Art} domain, and so on.

DomainNet \cite{DomainNet} is a large-scale dataset which contains 345 classes from six domains, while many classes contains mislabeled outliers and plenty of indistinguishable samples exist in domains \textit{Quickdraw} and \textit{Infograph} \cite{DomainNet40}. So Tan \textit{et al.} \cite{DomainNet40} select and construct a subset termed DomainNet40, which contains commonly-seen 40 classes from four domains: \textit{Clipart} (\textbf{C}), \textit{Painting} (\textbf{P}), \textit{Real} (\textbf{R}), and \textit{Sketch} (\textbf{S}). Similar to the OfficeHome, we split 20:20 classes for labeled and unlabeled data and establish twelve experiment settings for cross domain scenarios.

For motivation setup and warm-up experiments, we change the distribution of the unlabeled data by introducing several corruptions \cite{Corruptions} to create toy synthesized datasets. Further details have been mentioned in \ref{Motivation_Setup}. All of the experiments in this paper are conducted by training and testing models on corresponding datasets to ensure consistent experimental settings.

\textbf{Data Augmentation and Parameters Setup.}
We use strong data augmentation for all datasets, following the approach in UNO \cite{UNO} which includes crop, flip, and jittering in a moderate random manner. For a fair comparison, we also apply these transformations to all NCD methods thorough all of experiments.

The global hyper-parameters are set as follows: as for the baseline SimGCD \cite{SimGCD}, the temperatures $\tau_{u}$, $\tau_{c}$, $\tau_{s}$, and $\tau_{t}$ are set to 0.07, 1.0, 0.1 and 0.07 respectively. The balance parameter $\lambda$ in $\mathcal{L}_{\text{rep}}$ is 0.35, and $\varepsilon$ in $\mathcal{L}_{\mathrm{cls}}^{u}$ is set to 1.Besides, the parameters specific to the other methods RS \cite{RS}, SimGCD \cite{SimGCD}, UNO \cite{UNO}, and ComEx \cite{ComEx} are set to the same values as reported in original papers. With a batch size of 64, all of models are trained for 200 epochs using the SGD optimizer, with a weight decay of $5\times10^{-5}$ and momentum set to 0.9. The initial learning rate of 0.01 is decayed using cosine annealing, with a minimum value of $1\times10^{-5}$.The crop size of images is set to $32\times32$ for datasets related to CIFAR10 \cite{CIFAR10} and $128\times128$ for those sets regarding OfficeHome \cite{OfficeHome} and DomainNet40 \cite{DomainNet40}. All experiments are implemented using PyTorch and trained on a single NVIDIA RTX 4090 GPU, and results in our paper are the mean values with standard deviation (std) of 5 runs with different random seeds. In addition, the $w$ in the overall objective is set differently for each dataset and method to ensure the best performance, which will be mentioned in the corresponding sections.

\textbf{Evaluation Metrics.}
All of the experiments in this paper assess the performance of clustering on the \textit{novel} categories. As a primary evaluation metric used in clustering tasks, \textit{clustering ACCuracy} (ACC)\cite{ACC} is used here which is calculated as follows:
$$ {\rm ACC}=\frac{1}{N} \sum_{i=1}^{N} \mathbbm{1}\left[\boldsymbol{y}_{i}=map\left(\hat{\boldsymbol{y}}_{i}\right)\right],$$
where $\boldsymbol{y}_{i}$ and $\hat{\boldsymbol{y}}_{i}$ are ground-truth label and clustering assignment. $N$ is the number of the test data and the $map$ is the optimal permutation of predicted cluster indices computed via the Hungarian algorithm \cite{hungarian}.
Another common metric is the \textit{Normalized Mutual Information} (NMI) defined as:
$${\rm NMI}=\frac{MI\left(\boldsymbol{y}, \hat{\boldsymbol{y}}\right)}{\sqrt{H\left(\boldsymbol{y}\right) H\left(\hat{\boldsymbol{y}}\right)}},$$
where $MI\left(\boldsymbol{y},\hat{\boldsymbol{y}}\right)$ is the \textit{Mutual Information} between $\boldsymbol{y}$ and $\hat{\boldsymbol{y}}$, in which $\boldsymbol{y}$ is the set of ground-truth labels $\left\{\boldsymbol{y}_{i}\right\}_N$ and so on.

Besides ACC and NMI, \textit{Adjusted Rand Index} (ARI) is also used here to measure the agreement between clusters which is defined as:
$${\rm ARI}=\frac{RI - E(RI)}{\max(RI)-E(RI)},$$ 
where $RI = \frac{TP + TN}{TP + FP + FN + TN}$ is the \textit{Rand Index}, $E(RI)$ is the expected value of the $RI$ and $\max(RI) = 1$. $TP$, $TN$, $FP$, and $FN$ are the number of true positive, true negative, false positive, and false negative respectively. Different from ACC and NMI which range from 0 to 1, ARI ranges from -1 to 1. The value 0 indicates a random cluster and higher values indicate better clustering results.

\subsection{Warm-up Experiments and Benchmark Setup}
\label{Benchmark_Setup}

During the process of experiments, we observed that several NCD methods are built using different backbones, so we conducted a series of warm-up experiments to compare the performance of UNO \cite{UNO} and ComEx \cite{ComEx} with ResNet50, ResNet18 \cite{ResNet} and ViT \cite{ViT} backbones with different pre-trained strategies on both original and synthesized OfficeHome \cite{OfficeHome} datasets. The synthesized datasets were used to evaluate the performance in scenarios with distribution shifts between labeled and unlabeled sets while the original set is utilized to test algorithms with a real cross domain setting.

Specifically, similar to the toy dataset \textit{CIFAR10cmix}, the labeled data in the synthesized OfficeHome is based on the Real-World domain of the original set, while the unlabeled part is constructed using three corruption functions: gaussian blur, jpeg compression, and impulse noise (referred to as \textit{gaussian}, \textit{jpeg}, and \textit{impulse} respectively for simplicity) with a severity level of 5. Additionally, for consistency, the synthesized OfficeHome is denoted as \textit{OfficeHomecmix}. 
For the original OfficeHome, each model trained on the twelve cross domain settings has twelve corresponding testing results, and the mean value of the results with standard deviation (std) is listed for concise comparison.

The experimental results on metric ACC(\%) with std(\%) are presented in Tab. \ref{SOTA_change_backbone}. It could be observed that when using the same self-supervised training strategy \cite{DINO}, both UNO and ComEx perform significantly better with the ViT-b \cite{ViT} backbone compared to the counterparts with ResNet50 and ResNet18 \cite{ResNet}. Besides, it is evident that employing self-supervised pre-trained ResNet50 leads to improved performance for both methods compared to models without pre-training. However, training ResNet18 in the manner of DINO \cite{DINO} proved to be challenging, and the results with ResNet18 are unsatisfactory regardless of whether the backbone was pre-trained or not.

\begin{table}[htbp]
  \begin{center}
  \caption{Results of ACC(\%) with std(\%) of NCD methods with different backbones, with and without pre-training \cite{DINO}, on synthesized and original OfficeHome \cite{OfficeHome}. The best and second-best results are in bold and underlined respectively.}
  \label{SOTA_change_backbone}
  \resizebox{\linewidth}{!}{
  \begin{tabular}{c|lc|lll|l}
    \toprule
    ~ & ~ & ~ & \multicolumn{3}{c|}{OfficeHomecmix} & OfficeHome \\
    ~ & backbone & pre-train & gaussian & jpeg & impulse & \textbf{Mean} \\
    \midrule
    \multirow{5}{*}{UNO\cite{UNO}} & ViT-b & \ding{51} &
    $\mathbf{69.30}_{(1.4)}$  & $\mathbf{72.06}_{(1.2)}$ & $\mathbf{61.90}_{(1.4)}$ & $\mathbf{58.87}_{(1.8)}$ \\
    &ResNet50 & \ding{55} & $28.46_{(1.0)}$ & $29.12_{(0.8)}$ & $24.27_{(1.2)}$ & $28.25_{(1.0)}$ \\
    &ResNet50 & \ding{51} & $\underline{56.22}_{(1.4)}$ & $\underline{58.72}_{(1.2)}$ & $\underline{38.85}_{(1.0)}$ & $\underline{49.86}_{(1.6)}$ \\
    &ResNet18 & \ding{55}  & $30.91_{(1.2)}$ & $32.94_{(1.2)}$ & $29.95_{(1.5)}$ & $29.76_{(1.0)}$ \\
    &ResNet18 & \ding{51} & $30.08_{(0.8)}$ & $33.62_{(1.1)}$ & $29.35_{(1.1)}$  & $29.46_{(1.5)}$ \\
    \midrule
    \multirow{5}{*}{ComEx\cite{ComEx}}&ViT-b & \ding{51} &  $\mathbf{67.60}_{(1.6)}$ & $\mathbf{72.32}_{(1.7)}$ & $\mathbf{64.66}_{(2.2)}$ & $\mathbf{58.36}_{(2.4)}$ \\
    &ResNet50 & \ding{55} & $26.33_{(0.8)}$ & $30.52_{(0.8)}$ & $26.95_{(0.9)}$  & $28.15_{(1.0)}$ \\
    &ResNet50 & \ding{51} & $\underline{50.91}_{(3.1)}$ & $\underline{58.05}_{(1.5)}$ & $\underline{40.10}_{(0.6)}$  & $\underline{47.85}_{(1.7)}$ \\
    &ResNet18 & \ding{55} & $26.48_{(1.2)}$ & $33.93_{(1.2)}$ & $28.96_{(1.5)}$  & $29.29_{(0.9)}$ \\
    &ResNet18 & \ding{51} & $26.33_{(0.3)}$ & $33.75_{(0.8)}$ & $27.99_{(1.6)}$ & $29.34_{(1.3)}$ \\
    \bottomrule
  \end{tabular}}
  \end{center}
\end{table}

Based on these warm-up experiments, it is clear that the choice of backbone and pre-trained strategy plays a crucial role in the performance of NCD methods. Therefore, it is essential to establish a fair benchmark for comparison. In the following experiments, including in Sec. \ref{Motivation_Setup}, we use the ViT-b \cite{ViT} backbone and pre-trained manner \cite{DINO} for all NCD methods \cite{UNO, ComEx, SimGCD} to build a benchmark.

It is worth noting that representative one-stage NCD method RS \cite{RS} performs well using ResNet18 \cite{ResNet} with three successive steps: self-supervised training with all data; supervised training with labeled data; and finally auto-novel step using Rank Statistics to measure and match the similarities among unlabeled data points, subsequently facilitating the generation of pseudo labels. Although the training procedure is time-consuming, the compact and unified training strategies guarantee good results. When the pre-trained ViT-b \cite{ViT} is used as a backbone or Rank Statistics is used just for pseudo labeling in the supervised learning step, the performance of RS is consistently unsatisfactory. Therefore, we do not use RS in the following experiments, except for the Section of motivation setup \ref{Motivation_Setup} and call-back \ref{Motivation_Callback}, where we use the original RS for comparison.

\subsection{Novel Class Discovery Task on Toy Datasets with Distribution Shift}\label{experiments_on_toy_datasets}

\begin{table}[htbp]
  \begin{center}
  \caption{Results of ACC(\%) with std(\%) on two toy datasets with distribution shift. The best and second-best results are in bold and underlined respectively. The upper arrow indicates there exists improvement with the assistance of proposed module compared with the corresponding original methods.}
  \label{style_removing_acc}
  \resizebox{\linewidth}{!}{
  \begin{tabular}{l|lll|lll}
    \toprule
    ~ & \multicolumn{3}{c|}{CIFAR10cmix} & \multicolumn{3}{c}{OfficeHomecmix} \\
    ~ & gaussian & jpeg & impulse & gaussian & jpeg & impulse \\
    \midrule
    SimGCD\cite{SimGCD} & $53.35_{(3.3)}$ & $74.66_{(0.7)}$ & $76.00_{(0.4)}$ & $64.17_{(2.5)}$ & $67.81_{(4.1)}$ &  $57.60_{(3.5)}$ \\
    \midrule
    $+\mathcal{L}_{\mathrm{orth}}$ & $\mathbf{55.11}_{(0.4)}$\textuparrow & $76.79_{(3.0)}$\textuparrow &  $\underline{76.24}_{(0.2)}$\textuparrow & $66.04_{(1.8)}$\textuparrow & $68.23_{(3.1)}$\textuparrow & $58.86_{(1.5)}$\textuparrow\\
    $+\mathcal{L}_{\mathrm{cossimi}}$ & $\underline{54.69}_{(0.6)}$\textuparrow & $77.63_{(3.5)}$\textuparrow & $76.17_{(0.4)}$\textuparrow & $64.69_{(2.3)}$\textuparrow & $67.60_{(4.3)}$ & $58.02_{(3.3)}$\textuparrow\\
    $+\mathcal{L}_{\mathrm{corr}}$ & $54.06_{(1.7)}$\textuparrow & $78.17_{(2.3)}$\textuparrow & $\mathbf{76.38}_{(0.4)}$\textuparrow & $64.37_{(2.3)}$\textuparrow & $67.71_{(3.6)}$ & $58.12_{(4.1)}$\textuparrow\\
    \midrule
    \midrule
    UNO\cite{UNO} & $44.09_{(3.0)}$ & $77.84_{(0.4)}$ & $75.34_{(0.1)}$ & $69.30_{(1.4)}$ & $72.06_{(1.2)}$ & $61.90_{(1.4)}$ \\
    \midrule
    $+\mathcal{L}_{\mathrm{orth}}$ & $50.28_{(3.4)}$\textuparrow & $77.24_{(1.2)}$ & $75.38_{(0.1)}$\textuparrow & $69.14_{(0.7)}$ & $73.41_{(1.2)}$\textuparrow & $62.9_{(2.7)}$\textuparrow \\
    $+\mathcal{L}_{\mathrm{cossimi}}$ & $49.43_{(3.5)}$\textuparrow & $78.22_{(0.2)}$\textuparrow & $74.93_{(0.6)}$ & $\underline{71.77}_{(0.9)}$\textuparrow & $\underline{73.57}_{(0.6)}$\textuparrow & $62.6_{(2.8)}$\textuparrow \\
    $+\mathcal{L}_{\mathrm{corr}}$ & $49.40_{(3.5)}$\textuparrow & $78.25_{(0.2)}$\textuparrow & $74.90_{(0.6)}$ & $\mathbf{71.87}_{(1.0)}$\textuparrow & $\mathbf{73.70}_{(0.6)}$\textuparrow & $62.8_{(2.8)}$\textuparrow \\
    \midrule
    \midrule
    ComEx\cite{ComEx} & $41.63_{(3.3)}$ & $79.59_{(0.7)}$ & $64.91_{(4.4)}$ & $67.60_{(1.6)}$ & $72.32_{(1.7)}$ & $64.66_{(2.2)}$ \\
    \midrule
    $+\mathcal{L}_{\mathrm{orth}}$ & $44.33_{(3.6)}$\textuparrow & $\mathbf{80.43}_{(0.5)}$\textuparrow & $64.75_{(7.0)}$ & $68.72_{(2.8)}$\textuparrow & $72.73_{(2.1)}$\textuparrow & $65.36_{(2.5)}$\textuparrow \\
    $+\mathcal{L}_{\mathrm{cossimi}}$ & $42.18_{(2.6)}$\textuparrow & $80.02_{(0.6)}$\textuparrow & $66.14_{(6.7)}$\textuparrow & $68.02_{(2.0)}$\textuparrow & $72.66_{(1.8)}$\textuparrow & $\mathbf{66.95}_{(4.0)}$\textuparrow \\
    $+\mathcal{L}_{\mathrm{corr}}$ & $42.34_{(2.7)}$\textuparrow & $\underline{80.22}_{(0.6)}$\textuparrow & $64.62_{(8.4)}$ & $68.49_{(2.2)}$\textuparrow & $72.60_{(2.3)}$\textuparrow & $\underline{66.61}_{(4.8)}$\textuparrow \\
    \bottomrule
  \end{tabular}
  }
  \end{center}
\end{table}

In this part, we use synthesized toy datasets \textit{CIFAR10cmix} and \textit{OfficeHomecmix} mentioned in Sec. \ref{Motivation_Setup} and \ref{Benchmark_Setup} with distribution shift to verify our style removal module integrated with three baselines \cite{UNO, ComEx, SimGCD}. The baseline methods combined with our proposed module based on three style removal objective functions \ref{L_orth}, \ref{L_cossimi} and \ref{L_corr}, is respectively denoted as $+\mathcal{L}_{\mathrm{orth}}$, $+\mathcal{L}_{\mathrm{cossimi}}$ and $+\mathcal{L}_{\mathrm{corr}}$. The trade-off parameter $w$ in total objective of improved SimGCD \cite{SimGCD} is set to 0.01 for \textit{CIFAR10cmix} and 0.05 for \textit{OfficeHomecmix} in toy experiments. Besides, $w$ is set to 0.01 for both UNO \cite{UNO} and ComEx \cite{ComEx} counterparts.

From the results shown in Tab. \ref{style_removing_acc}, it can obviously be seen that with the assistance of style removal module, SimGCD \cite{SimGCD} has been improved to some extent on both \textit{CIFAR10cmix} and \textit{OfficeHomecmix} datasets with different corruptions and the improvement is significant especially on \textit{CIFAR10cmix}. Regarding the OfficeHome dataset with higher resolution, UNO \cite{UNO} and ComEx \cite{ComEx} achieve better results using a multi-view self-labeling strategy and combining the style removal module brings general improvement.

Moreover, the parameter $w$ is set to 0.001, 0.005, 0.01, 0.05, and 0.1 to find the optimal result on the validation sets. Except for the case when the $w$ is set to 0.1 where the model struggles to converge, the results for the other values show no significant differences. To sum up, the experiments on toy datasets with distribution shift show that the performance of our module is not sensitive to the choice of different style removal objective functions and corresponding trade-off parameter, confirming the robustness of the proposed module.

\subsection{Cross Domain Novel Class Discovery Task}

\begin{table*}[htbp]
  \begin{center}
  \caption{Results of ACC(\%) with std(\%) on OfficeHome \cite{OfficeHome}.}
  \label{cross_domain_Officehome_ACC}
  \resizebox{\linewidth}{!}{
  \begin{tabular}{l|lll|lll|lll|lll|l}
    \toprule
    & $\textbf{R}\rightarrow\textbf{A}$ & $\textbf{R}\rightarrow\textbf{C}$ & $\textbf{R}\rightarrow\textbf{P}$ & $\textbf{A}\rightarrow\textbf{R}$ & $\textbf{A}\rightarrow\textbf{C}$ & $\textbf{A}\rightarrow\textbf{P}$ & $\textbf{C}\rightarrow\textbf{R}$ & $\textbf{C}\rightarrow\textbf{A}$ & $\textbf{C}\rightarrow\textbf{P}$ & $\textbf{P}\rightarrow\textbf{R}$ & $\textbf{P}\rightarrow\textbf{A}$ & $\textbf{P}\rightarrow\textbf{C}$ & \textbf{Mean}\\
    \midrule
    SimGCD\cite{SimGCD} & $57.34_{(3.5)}$ & $39.79_{(3.1)}$ & $70.63_{(3.8)}$ & $68.75_{(2.7)}$ & $44.27_{(3.3)}$ & $\underline{70.55}_{(2.8)}$ & $72.71_{(1.8)}$ & $55.78_{(3.6)}$ & $70.16_{(4.6)}$ & $69.90_{(2.4)}$ & $58.75_{(1.9)}$ & $38.12_{(2.5)}$ & $59.73_{(3.0)}$\\
    \midrule
    $+\mathcal{L}_{\mathrm{orth}}$ & $57.66_{(2.8)}$\textuparrow & $40.10_{(2.7)}$\textuparrow & $\mathbf{71.56}_{(1.2)}$\textuparrow & $68.85_{(1.8)}$\textuparrow & $44.79_{(3.6)}$\textuparrow & $70.55_{(2.9)}$ & $72.81_{(2.4)}$\textuparrow & $\mathbf{58.91}_{(3.9)}$\textuparrow & $\mathbf{71.64}_{(3.1)}$\textuparrow & $70.73_{(2.7)}$\textuparrow & $58.91_{(2.4)}$\textuparrow & $40.10_{(3.0)}$\textuparrow & $\mathbf{60.55}_{(2.7)}$\textuparrow\\
    $+\mathcal{L}_{\mathrm{cossimi}}$ & $\mathbf{58.44}_{(2.7)}$\textuparrow & $39.69_{(3.7)}$ & $\underline{70.71}_{(3.7)}$\textuparrow & $68.96_{(3.1)}$\textuparrow & $44.37_{(3.9)}$\textuparrow & $70.39_{(3.3)}$ & $\mathbf{73.23}_{(2.1)}$\textuparrow & $56.87_{(3.6)}$\textuparrow & $\underline{71.41}_{(3.9)}$\textuparrow & $70.21_{(2.4)}$\textuparrow & $\mathbf{59.84}_{(3.1)}$\textuparrow & $40.11_{(2.7)}$\textuparrow & $\underline{60.35}_{(3.2)}$\textuparrow\\
    $+\mathcal{L}_{\mathrm{corr}}$ & $\underline{58.28}_{(2.4)}$\textuparrow & $39.58_{(3.5)}$ & $70.63_{(3.8)}$\textuparrow & $69.06_{(3.1)}$\textuparrow & $44.58_{(3.6)}$\textuparrow & $\mathbf{70.71}_{(4.2)}$\textuparrow & $\underline{73.23}_{(2.3)}$\textuparrow & $\underline{56.88}_{(3.7)}$\textuparrow & $70.31_{(5.3)}$\textuparrow & $70.21_{(2.8)}$\textuparrow & $\underline{59.37}_{(2.4)}$\textuparrow & $39.38_{(3.0)}$\textuparrow & $60.19_{(3.3)}$\textuparrow\\
    \midrule
    \midrule
    UNO \cite{UNO} & $51.76_{(1.5)}$ & $\mathbf{44.35}_{(1.9)}$ & $67.21_{(3.1)}$ & $72.79_{(1.4)}$ & $45.94_{(2.2)}$ & $68.36_{(1.4)}$ & $70.68_{(1.9)}$ & $51.99_{(2.8)}$ & $65.10_{(1.7)}$ & $71.67_{(1.1)}$ & $52.73_{(1.7)}$ & $43.91_{(1.3)}$ & $58.87_{(1.8)}$ \\
    \midrule
    $+\mathcal{L}_{\mathrm{orth}}$ & $52.15_{(0.3)}$\textuparrow & $\underline{44.27}_{(1.9)}$ & $66.50_{(2.3)}$ & $\underline{73.10}_{(2.1)}$\textuparrow & $46.72_{(2.2)}$\textuparrow & $68.03_{(2.5)}$ & $68.98_{(2.0)}$ & $52.77_{(0.9)}$\textuparrow & $65.45_{(0.3)}$\textuparrow &$71.93_{(1.1)}$\textuparrow & $51.91_{(0.7)}$ & $43.80_{(1.3)}$ & $58.80_{(1.5)}$ \\
    $+\mathcal{L}_{\mathrm{cossimi}}$ & $52.58_{(2.2)}$\textuparrow & $43.98_{(2.5)}$ & $67.34_{(3.2)}$\textuparrow & $72.50_{(1.9)}$ & $46.15_{(2.0)}$\textuparrow & $68.06_{(0.9)}$ & $70.81_{(2.0)}$\textuparrow & $53.05_{(1.6)}$\textuparrow & $65.39_{(1.5)}$\textuparrow & $71.98_{(1.5)}$\textuparrow & $52.42_{(1.5)}$ & $44.24_{(1.3)}$\textuparrow & $59.04_{(1.8)}$\textuparrow \\
    $+\mathcal{L}_{\mathrm{corr}}$ & $52.66_{(1.7)}$\textuparrow & $43.93_{(2.3)}$ & $67.32_{(3.2)}$\textuparrow & $72.71_{(1.7)}$ & $46.04_{(1.9)}$\textuparrow & $68.14_{(0.8)}$ & $70.94_{(2.3)}$\textuparrow & $53.05_{(1.9)}$\textuparrow & $65.37_{(1.5)}$\textuparrow & $71.56_{(1.4)}$ & $52.50_{(1.5)}$ & $44.32_{(1.2)}$\textuparrow & $59.04_{(1.8)}$\textuparrow \\
    \midrule
    \midrule
    ComEx \cite{ComEx} & $49.22_{(1.0)}$ & $43.44_{(1.9)}$ & $65.86_{(1.7)}$ & $72.24_{(2.3)}$ & $46.90_{(2.8)}$ & $68.89_{(2.1)}$ & $70.76_{(4.9)}$ & $49.37_{(2.4)}$ & $66.68_{(3.0)}$ & $\mathbf{73.41}_{(1.4)}$ & $47.93_{(3.1)}$ & $45.68_{(1.7)}$ & $58.36_{(2.4)}$ \\
    \midrule
    $+\mathcal{L}_{\mathrm{orth}}$ & $48.79_{(2.0)}$ & $43.70_{(1.0)}$\textuparrow & $66.37_{(2.9)}$\textuparrow & $\mathbf{73.39}_{(2.8)}$\textuparrow & $46.25_{(1.5)}$ & $68.40_{(3.1)}$ & $70.96_{(4.2)}$\textuparrow & $50.39_{(0.8)}$\textuparrow & $67.54_{(3.0)}$\textuparrow & $72.71_{(2.0)}$ & $48.0_{(2.2)}$\textuparrow & $44.53_{(1.2)}$ & $58.42_{(2.2)}$\textuparrow\\
    $+\mathcal{L}_{\mathrm{cossimi}}$ & $48.87_{(1.2)}$ & $44.06_{(2.2)}$\textuparrow & $65.70_{(2.1)}$ & $72.06_{(2.4)}$ & $\underline{47.19}_{(1.7)}$\textuparrow & $69.33_{(0.7)}$\textuparrow & $71.67_{(3.0)}$\textuparrow & $49.26_{(1.1)}$ & $66.11_{(3.1)}$ & $\underline{73.1}_{(2.2)}$ & $48.2_{(2.4)}$\textuparrow & $\mathbf{46.77}_{(2.8)}$\textuparrow & $58.53_{(2.1)}$\textuparrow\\
    $+\mathcal{L}_{\mathrm{corr}}$ & $49.37_{(1.1)}$\textuparrow & $43.57_{(1.6)}$\textuparrow & $65.76_{(1.6)}$ & $72.92_{(2.4)}$\textuparrow & $\mathbf{47.42}_{(1.7)}$\textuparrow & $69.55_{(1.0)}$\textuparrow & $71.78_{(3.6)}$\textuparrow & $49.14_{(1.5)}$ & $65.86_{(2.6)}$ & $72.0_{(2.2)}$ & $47.5_{(2.6)}$ & $\underline{46.64}_{(1.7)}$\textuparrow & $58.48_{(2.0)}$\textuparrow \\
    \bottomrule
  \end{tabular}}
  \end{center}
\end{table*}

In the realistic cross domain situation, similar to the warm-up part in Sec. \ref{Benchmark_Setup}, we employ the twelve scenarios with different domains between labeled and unlabeled sets to sequentially train and test the baseline NCD methods and their improved counterparts integrated with proposed module. The performance results on OfficeHome \cite{OfficeHome} and DomainNet40 \cite{DomainNet40} evaluated by metric ACC are shown in Tab. \ref{cross_domain_Officehome_ACC} and Tab. \ref{cross_domain_DomainNet40_ACC} respectively. Due to the space limitation, the results of NMI and ARI are provided in \ref{NMI_ARI_test_results} in Appendix. The $w$ in the total objective of SimGCD \cite{SimGCD} is set to 0.05 for OfficeHome and 0.005 for DomainNet40 in cross domain experiments. Meanwhile, all $w$ is set to 0.01 for both UNO \cite{UNO} and ComEx \cite{ComEx} on two datasets.

\begin{table*}[htbp]
  \begin{center}
  \caption{Results of ACC(\%) with std(\%) on DomainNet40 \cite{DomainNet40}.}
  \label{cross_domain_DomainNet40_ACC}
  \resizebox{\linewidth}{!}{
  \begin{tabular}{l|lll|lll|lll|lll|l}
    \toprule
    & $\textbf{R}\rightarrow\textbf{C}$ & $\textbf{R}\rightarrow\textbf{P}$ & $\textbf{R}\rightarrow\textbf{S}$ & $\textbf{C}\rightarrow\textbf{R}$ & $\textbf{C}\rightarrow\textbf{P}$ & $\textbf{C}\rightarrow\textbf{S}$ & $\textbf{P}\rightarrow\textbf{R}$ & $\textbf{P}\rightarrow\textbf{C}$ & $\textbf{P}\rightarrow\textbf{S}$ & $\textbf{S}\rightarrow\textbf{R}$ & $\textbf{S}\rightarrow\textbf{C}$ & $\textbf{S}\rightarrow\textbf{P}$ & \textbf{Mean} \\
    \midrule
    SimGCD\cite{SimGCD} & $50.17_{(1.8)}$ & $58.65_{(2.4)}$ & $39.15_{(2.4)}$ & $81.24_{(1.4)}$ & $61.94_{(3.9)}$ & $39.84_{(2.6)}$ & $81.53_{(2.1)}$ & $49.42_{(3.0)}$ & $39.65_{(1.1)}$ & $83.80_{(1.4)}$ & $51.17_{(3.8)}$ & $64.04_{(3.5)}$ & $58.38_{(2.5)}$ \\
    \midrule
    $+\mathcal{L}_{\mathrm{orth}}$ & $50.17_{(2.0)}$ & $59.15_{(3.7)}$\textuparrow & $39.39_{(0.9)}$\textuparrow & $81.95_{(1.9)}$\textuparrow & $\mathbf{62.72}_{(3.4)}$\textuparrow & $40.34_{(2.5)}$\textuparrow & $82.33_{(2.6)}$\textuparrow & $50.27_{(3.8)}$\textuparrow & $40.03_{(1.1)}$\textuparrow & $83.90_{(1.2)}$\textuparrow & $51.23_{(4.7)}$\textuparrow & $\mathbf{64.35}_{(3.2)}$\textuparrow & $58.82_{(2.6)}$\textuparrow \\
    $+\mathcal{L}_{\mathrm{cossimi}}$ & $49.39_{(1.7)}$ & $59.23_{(3.6)}$\textuparrow & $38.72_{(2.3)}$ & $81.19_{(1.5)}$ & $62.07_{(3.8)}$\textuparrow & $39.87_{(2.9)}$\textuparrow & $81.74_{(2.0)}$\textuparrow & $49.29_{(3.0)}$ & $39.74_{(0.9)}$\textuparrow & $83.94_{(1.2)}$\textuparrow & $51.37_{(3.9)}$\textuparrow & $64.04_{(3.7)}$ & $58.38_{(2.6)}$ \\
    $+\mathcal{L}_{\mathrm{corr}}$ & $49.27_{(1.7)}$ & $59.15_{(3.8)}$\textuparrow & $39.05_{(1.9)}$ & $81.66_{(1.4)}$\textuparrow & $\underline{62.10}_{(3.9)}$\textuparrow & $39.97_{(2.6)}$\textuparrow & $81.73_{(2.0)}$\textuparrow & $49.75_{(2.5)}$\textuparrow & $39.76_{(1.2)}$\textuparrow & $83.99_{(1.3)}$\textuparrow & $51.15_{(3.7)}$ & $\underline{64.11}_{(3.5)}$\textuparrow & $58.47_{(2.5)}$\textuparrow \\
    \midrule
    \midrule
    UNO \cite{UNO} & $52.69_{(2.1)}$ & $52.54_{(2.0)}$ & $45.34_{(1.5)}$ & $86.46_{(1.8)}$ & $53.85_{(1.3)}$ & $47.75_{(1.3)}$ & $86.03_{(1.3)}$ & $54.11_{(2.3)}$ & $46.05_{(0.8)}$ & $84.54_{(2.0)}$ & $56.50_{(1.8)}$ & $55.10_{(1.2)}$ & $60.08_{(1.6)}$ \\
    \midrule
    $+\mathcal{L}_{\mathrm{orth}}$ & $52.21_{(1.7)}$ & $53.14_{(2.2)}$\textuparrow & $44.71_{(2.0)}$ & $85.66_{(2.2)}$ & $54.13_{(1.3)}$\textuparrow & $46.85_{(1.7)}$ & $85.14_{(1.0)}$ & $\mathbf{54.59}_{(1.9)}$\textuparrow & $45.54_{(1.8)}$ & $83.80_{(1.7)}$ & $55.76_{(1.7)}$ & $54.53_{(0.9)}$ & $59.67_{(1.7)}$ \\
    $+\mathcal{L}_{\mathrm{cossimi}}$ & $\underline{53.29}_{(1.8)}$\textuparrow & $52.39_{(2.1)}$ & $\underline{45.88}_{(1.2)}$\textuparrow & $86.35_{(1.9)}$ & $53.55_{(1.2)}$ & $47.87_{(1.1)}$\textuparrow & $86.24_{(0.9)}$\textuparrow & $54.50_{(1.6)}$\textuparrow & $\underline{46.27}_{(1.2)}$\textuparrow & $84.90_{(2.4)}$\textuparrow & $\mathbf{56.81}_{(1.6)}$\textuparrow & $55.26_{(1.3)}$\textuparrow & $60.28_{(1.5)}$\textuparrow \\
    $+\mathcal{L}_{\mathrm{corr}}$ & $\mathbf{53.69}_{(1.5)}$\textuparrow & $52.56_{(1.8)}$\textuparrow & $\mathbf{46.09}_{(1.2)}$\textuparrow & $86.33_{(1.9)}$ & $53.48_{(1.1)}$ & $\mathbf{47.99}_{(1.0)}$\textuparrow & $86.25_{(0.9)}$\textuparrow & $\underline{54.56}_{(1.7)}$\textuparrow & $\mathbf{46.27}_{(1.0)}$\textuparrow & $84.93_{(2.4)}$\textuparrow & $\underline{56.80}_{(1.6)}$\textuparrow & $55.20_{(1.3)}$\textuparrow & $60.34_{(1.5)}$\textuparrow \\
    \midrule
    \midrule
    ComEx \cite{ComEx} & $52.30_{(1.6)}$ & $60.96_{(1.0)}$ & $43.77_{(2.1)}$ & $87.49_{(2.7)}$ & $61.69_{(1.7)}$ & $47.22_{(2.0)}$ & $\underline{87.32}_{(2.5)}$ & $52.11_{(1.7)}$ & $45.36_{(1.4)}$ & $86.18_{(2.9)}$ & $52.67_{(2.1)}$ & $61.99_{(1.8)}$ & $61.59_{(2.0)}$ \\
    \midrule
    $+\mathcal{L}_{\mathrm{orth}}$ & $50.90_{(1.7)}$ & $60.37_{(2.7)}$ & $44.88_{(2.4)}$\textuparrow & $\underline{87.6}_{(3.8)}$\textuparrow & $61.44_{(1.7)}$ & $47.68_{(1.4)}$\textuparrow & $\mathbf{87.41}_{(1.5)}$\textuparrow & $51.86_{(1.0)}$ & $44.81_{(3.1)}$ & $85.95_{(2.7)}$ & $53.02_{(2.9)}$\textuparrow & $62.17_{(0.7)}$\textuparrow & $61.51_{(2.1)}$ \\
    $+\mathcal{L}_{\mathrm{cossimi}}$ & $51.45_{(1.6)}$ & $\mathbf{61.69}_{(1.0)}$\textuparrow & $44.53_{(2.9)}$\textuparrow & $\mathbf{87.83}_{(2.9)}$\textuparrow & $61.26_{(1.1)}$ & $\underline{47.91}_{(1.5)}$\textuparrow & $86.25_{(1.9)}$ & $53.23_{(1.5)}$\textuparrow & $45.40_{(2.1)}$\textuparrow & $\mathbf{87.62}_{(1.5)}$\textuparrow & $53.03_{(1.3)}$\textuparrow & $61.75_{(1.8)}$ & $\mathbf{61.83}_{(1.8)}$\textuparrow \\
    $+\mathcal{L}_{\mathrm{corr}}$ & $52.0_{(2.0)}$ & $\underline{61.49}_{(0.8)}$\textuparrow & $44.12_{(3.4)}$\textuparrow & $87.35_{(2.3)}$ & $61.21_{(0.7)}$ & $47.64_{(1.5)}$\textuparrow & $86.47_{(1.7)}$ & $53.59_{(1.0)}$\textuparrow & $45.6_{(1.2)}$\textuparrow & $\underline{86.90}_{(1.5)}$\textuparrow & $52.77_{(1.6)}$\textuparrow & $61.94_{(1.8)}$ & $\underline{61.76}_{(1.6)}$\textuparrow \\
    \bottomrule
  \end{tabular}}
  \end{center}
\end{table*}
From the mean results shown in above tables, it could be concluded that on OfficeHome with cross domain setting, our module integrated with SimGCD \cite{SimGCD} generally outperforms the NCD SOTAs \cite{UNO} \cite{ComEx} on all three metrics. The upward arrows indicate that with the assistance of the proposed exclusive style removal module, which includes \textit{style encoder} $g$ and three different objectives $\mathcal{L}_{\mathrm{style\_removal}}$, the test results of pluged method on both datasets are better than corresponding baselines to varying extents in most of the cross domain conditions.

Besides, on both datasets, the SimGCD \cite{SimGCD} with $\mathcal{L}_{\mathrm{orth}}$ induced by simplest inner product generally performs the best compared to the model using $\mathcal{L}_{\mathrm{cossimi}}$ and $\mathcal{L}_{\mathrm{corr}}$. Additionally, the mean results in the last row indicate the same conclusion. This is consistent with the results shown in Table. \ref{style_removing_acc} on toy datasets and verifies that keeping the style and content features orthogonal is the most effective way to decouple the two kinds of features in this baseline.

However, when it comes to UNO \cite{UNO} and ComEx \cite{ComEx}, the improvement brought by $\mathcal{L}_{\mathrm{orth}}$ was not obvious, and it did not perform as well as $\mathcal{L}_{\mathrm{cossimi}}$ and $\mathcal{L}_{\mathrm{corr}}$ in both DomainNet40 \cite{DomainNet40} and OfficeHome \cite{OfficeHome}. The inconsistency with the toy datasets could be caused by the different impacts of different distribution shifts on the style features of samples, because the data distribution shifts brought by the corruption function and cross-domain are actually different. It is reasonable that diverse style remove functions have different improvement on the model for different distribution shifts.

Last but not least, it can be observed that the performance of each algorithm varies significantly with different matches between the source and target domains on two datasets. Additionally, compared to different source domains, the results seem to be more related to the target domain. Particularly when the target domain is \textit{real} of both OfficeHome \cite{OfficeHome} and DomainNet40 \cite{DomainNet40}, all algorithms perform very well. This is because the backbone is pre-trained on ImageNet, which can be referred as a dataset in \textit{real} domain, further highlighting the importance of pre-training.

\subsection{Motivation Call-back and Plug-in Ability of Proposed Module}
\label{Motivation_Callback}
When we revisit the motivation setup in Sec. \ref{Motivation_Setup}, our initial goal was to bridge the gap between two types of experimental settings: those with distribution shift between labeled and unlabeled sets, and those without. In other words, the degradation of performance on data with different distributions needs to be alleviated to some extent. To achieve this, we integrate the proposed exclusive style removal module into these NCD approaches. Similar to Sec. \ref{Motivation_Setup}, these three methods are trained and tested on the synthesized \textit{CIFAR10cmix} with different corruption severities of Gaussian Blur \cite{Corruptions} on \textit{novel} categories. And the simplest objective $\mathcal{L}_{\mathrm{orth}}$ in $\mathcal{L}_{\mathrm{style\_removal}}$ is added to the overall loss function of corresponding methods. Here the parameter $w$ of the $\mathcal{L}_{\mathrm{style\_removal}}$ is set to 0.01 for all methods.

\begin{figure}[htbp]
  \centering
  \subfigure[UNO\cite{UNO}]{
  \includegraphics[width=0.466\linewidth]{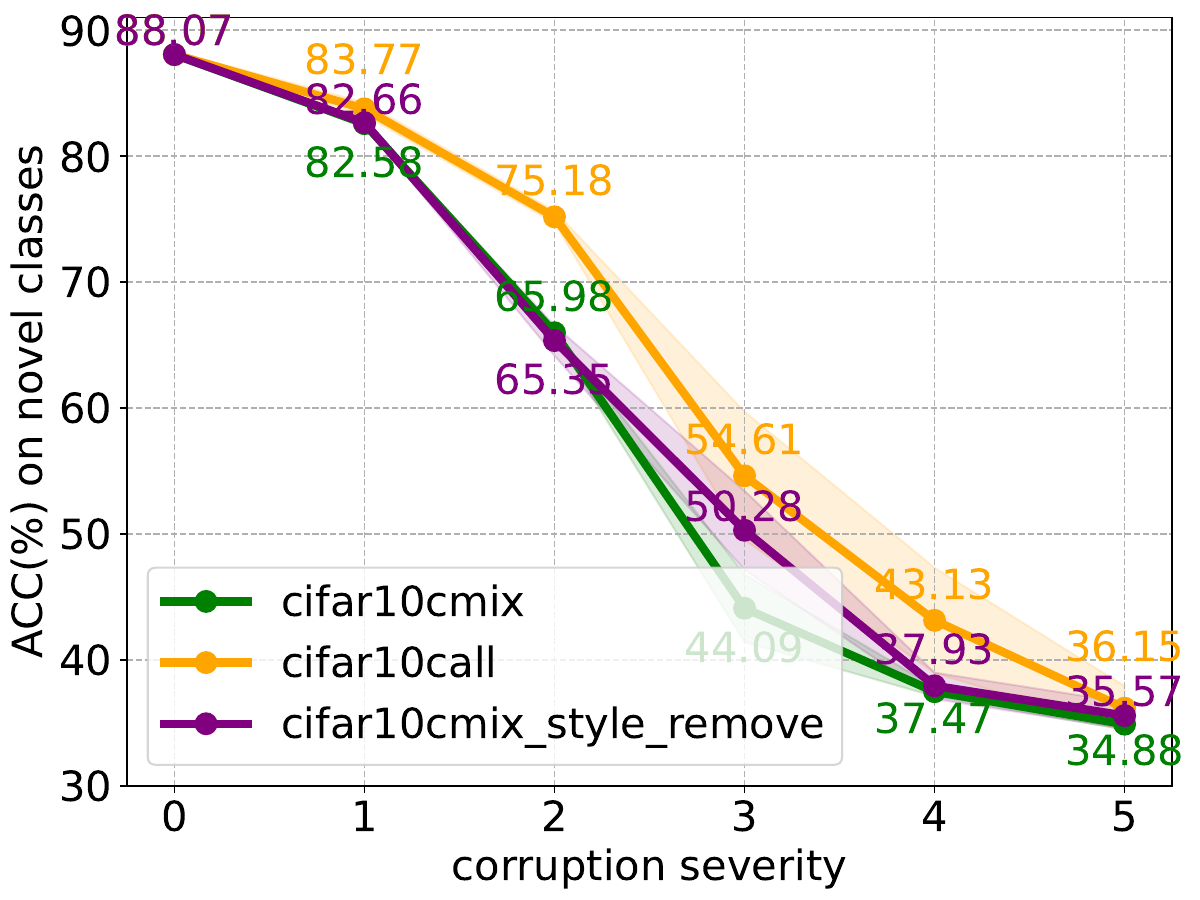}}
  \subfigure[ComEx\cite{ComEx}]{
  \includegraphics[width=0.466\linewidth]{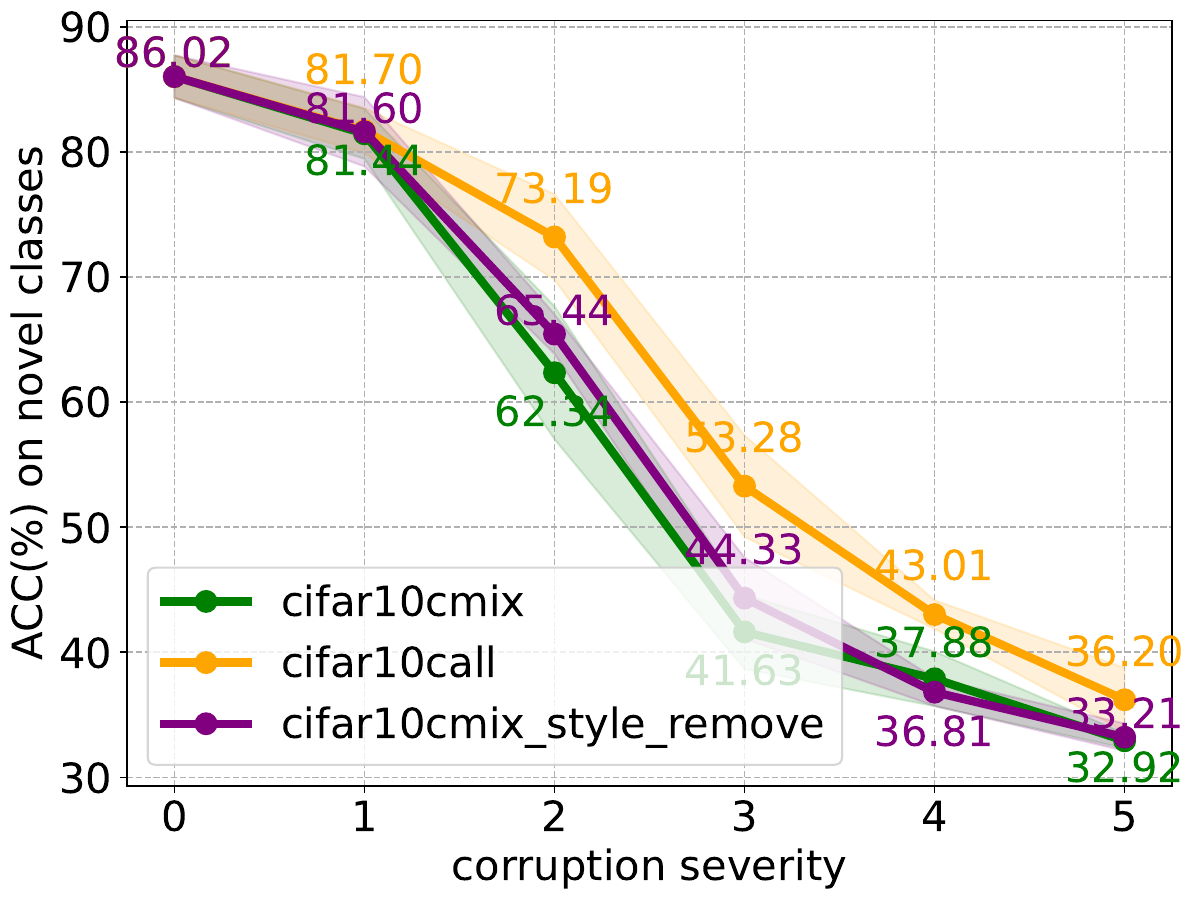}}
  \subfigure[RS\cite{RS}]{
  \includegraphics[width=0.466\linewidth]{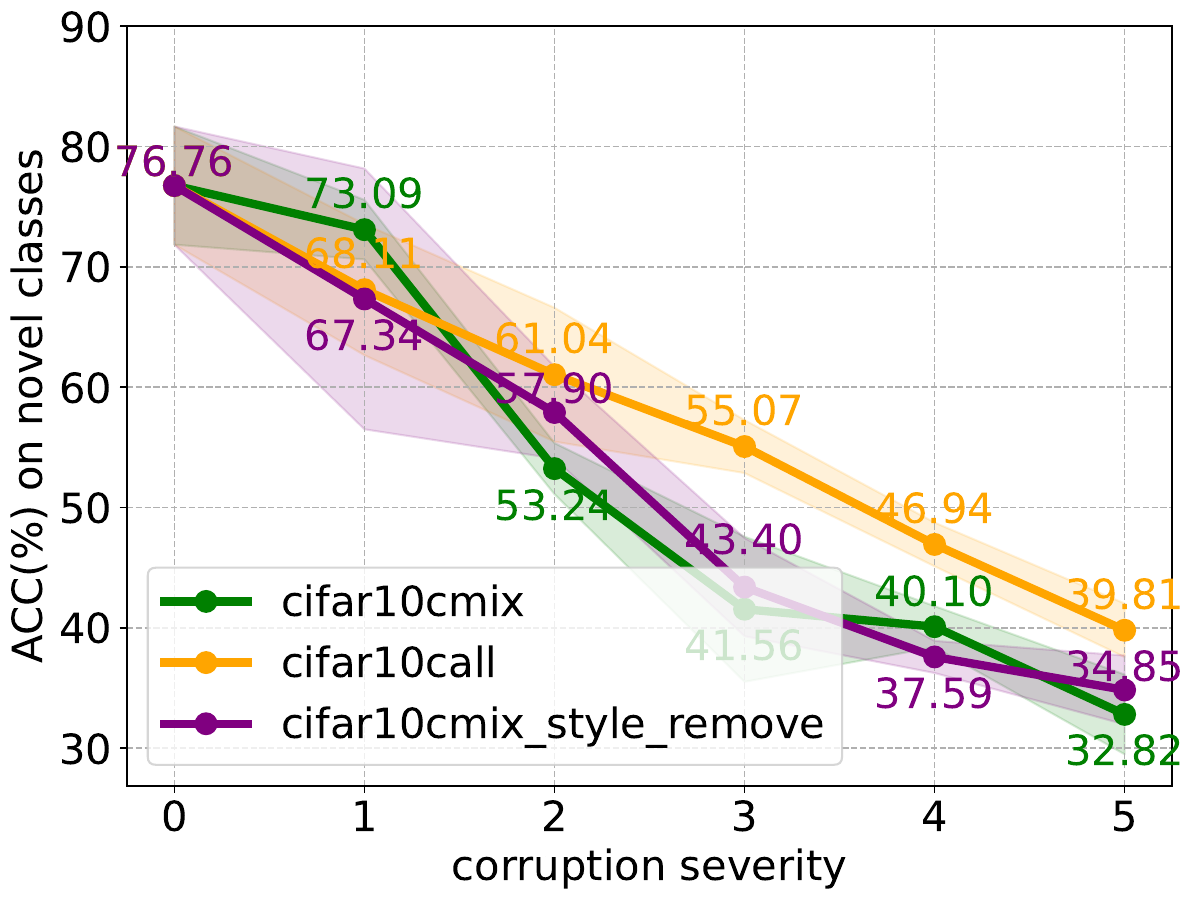}}
  \subfigure[Baseline\cite{SimGCD}]{
  \includegraphics[width=0.466\linewidth]{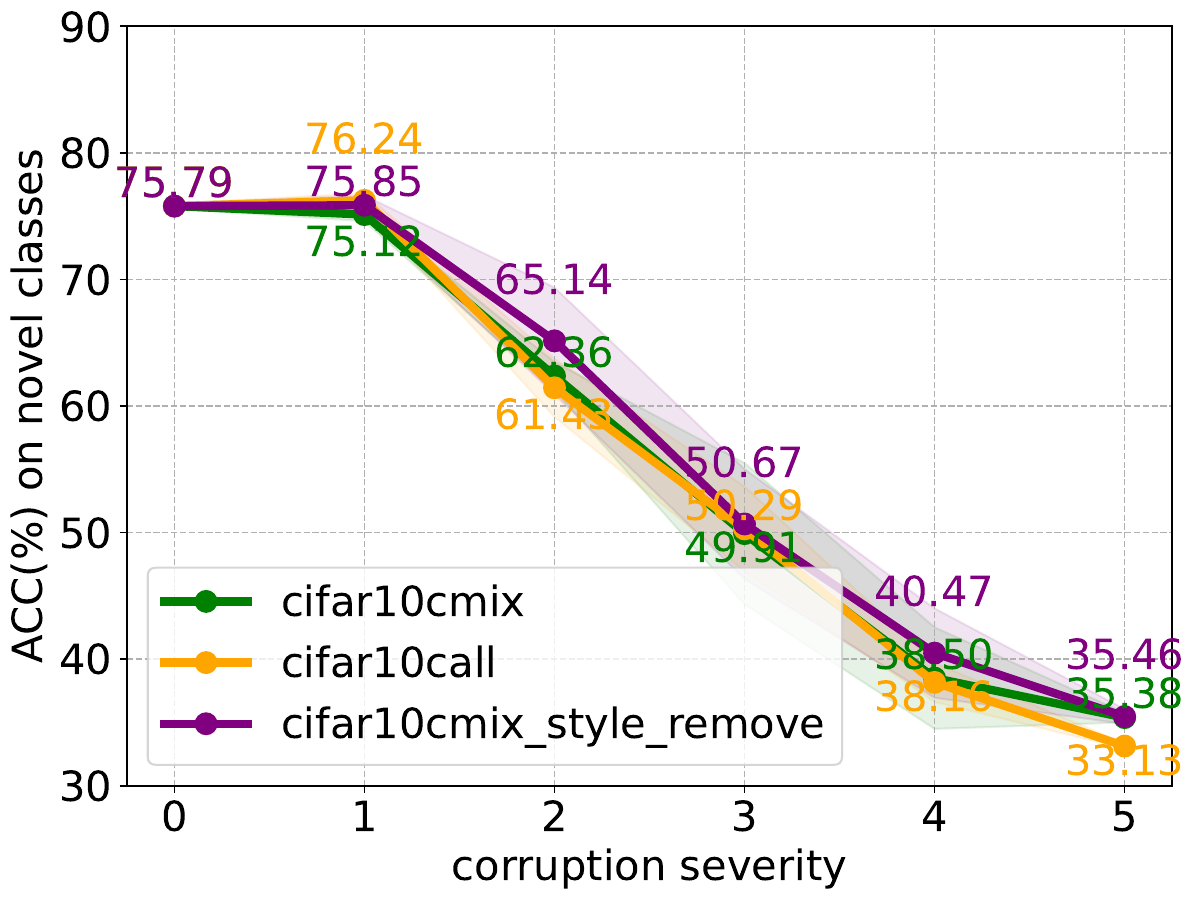}}
  \caption{The gap is bridged to some extent with the assistance of the proposed style removal module on existing NCD methods.}
  \label{motivation_callback}
\end{figure}

A series of test results are shown with newly added purple lines in Fig. \ref{motivation_callback} compared with Fig. \ref{motivation_setup}. From the results, we can see that with the assistance of the style removal module, the performance of these three methods is improved on synthesized \textit{CIFAR10cmix} and the gap becomes smaller to some extent than before shown in Fig. \ref{motivation_setup}. This confirms the effectiveness and the plug-in ability of the proposed module. 

\begin{figure}[htbp]
  \centering
  \subfigure[ComEx\cite{ComEx} (52.17)]{
  \includegraphics[width=0.463\linewidth]{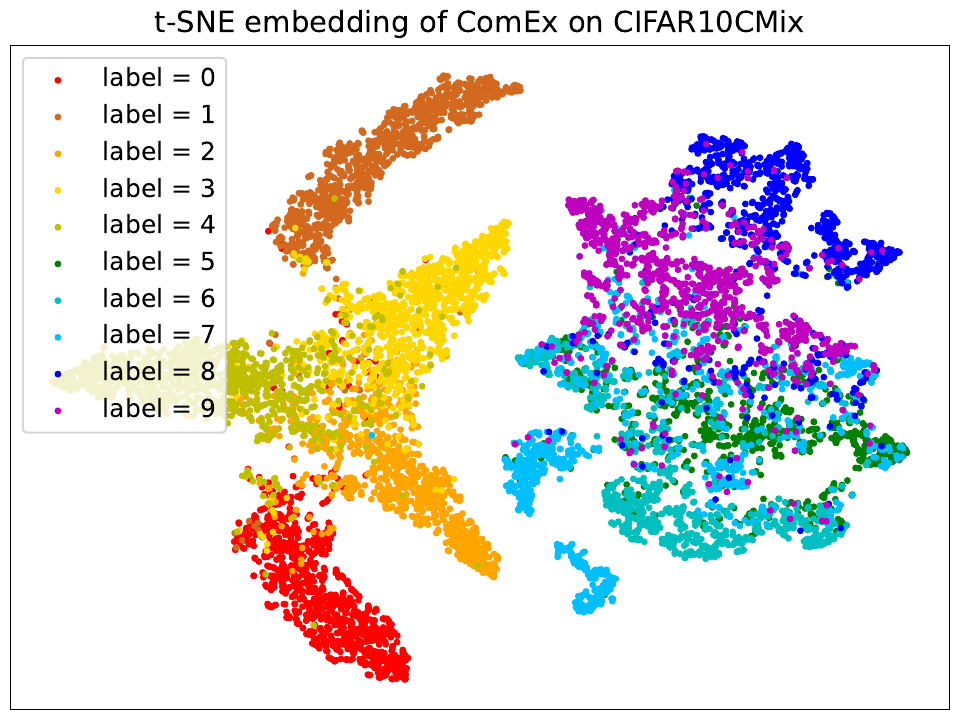}
  }
  \subfigure[ComEx$+\mathcal{L}_{\mathrm{orth}}$ (67.57)]{
  \includegraphics[width=0.463\linewidth]{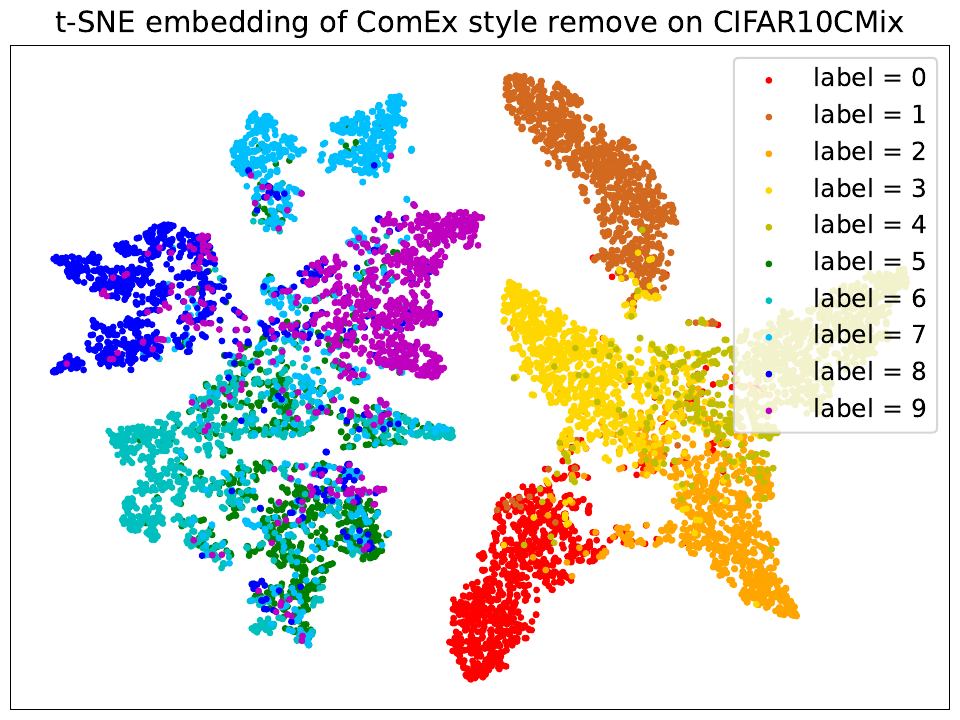}
  }
  \caption{The t-SNE embedding of the feature processed by backbones with and without proposed style removal module on synthesized \textit{CIFAR10cmix}. The values in the brackets are the cluster ACC(\%) on \textit{novel} categories.}
  \label{t-SNE}
\end{figure}

Meanwhile, it is also interesting to note that there is almost no gap between the two settings in the SimGCD \cite{SimGCD}, which may be due to the fact that with a contractive learning strategy, this method could learn more discriminative content features than RS \cite{RS}, UNO \cite{UNO} and ComEx \cite{ComEx}. In addition, when integrated with the style removal module, the style feature is removed more thoroughly, which makes the results even outperform the counterpart on datasets without distribution shift, as shown with an orange line in Fig. \ref{motivation_callback}(d).
In order to illustrate the improvement more clearly, we provide the t-SNE figures in Fig. \ref{t-SNE} showing the feature distribution obtained by \cite{ComEx} with and without the proposed module. Specifically, the feature fed for the t-SNE embedding method is the output of the backbone on one random seed, and the labels are the ground truth of the test set. The corruption level of Gaussian Blur on \textit{novel} categories is set to 2 and the $w$ is set to 0.01.

It is clear that for the last five classes (\textit{novel} categories), the feature distribution with the proposed style removal module is more compact and separable than vanilla ones, leading to higher cluster accuracy and a higher position of purple lines compared with green lines in Fig. \ref{motivation_callback}. Even for the first five classes (\textit{seen} categories), the method with the proposed module also shows a more compact feature distribution than the origin, which verifies the merit of the proposed module and its effectiveness in addressing the motivation.

\section{Conclusion}
In this paper, we introduce a task named \textit{Cross Domain Novel Class Discovery} and discuss its solvability. Based on the theoretical analysis, a new solution is proposed which utilizes a GCD algorithm as a baseline and includes a simple yet effective exclusive style removal module trained simultaneously with the baseline. Moreover, the proposed module can be easily integrated into several NCD methods as a plug-in to improve their performance on data with distribution shift. Experimental results show that these NCD methods integrated with proposed module method generally outperform the original ones on synthesized toy datasets with distribution shift and two common datasets with cross domain settings. Last but not least, a fair benchmark with the same backbone and pre-trained strategy built in this paper is beneficial for the development of NCD and other related transfer learning tasks.

{
\bibliographystyle{ieeetr}
\bibliography{citationlist}
}

\newpage
\section*{appendix}
\subsection{Proof of Dimension Lemma regarding $K$-$\epsilon$-separable r.v.}\label{proof_lemma}
\begin{proof}
  \textbf{Sufficiency:} We aim to prove that based on the same support set $R_{X \mid f(X)=i}$, that a $m$-dimension r.v. $Z \in \mathcal{Z} \subset \mathcal{W} \subset \mathcal{X}$ is $K$-$\epsilon$-separable is a sufficient condition for that a expanded $n$-dimension r.v. $W \in \mathcal{W}$ is $K$-$\epsilon$-separable.
  
  As $d$-dimension space $\mathcal{X}$ is bounded, the $n$-dimension subspace $\mathcal{W}$ is bounded and compact. Then for $\forall i \in \mathcal{I}$, there exists a limited number (set as $c_i$) of $n$-dimension open spares with diameter 1 can cover the support set $R_{W \mid f(X)=i}$. As index set $\mathcal{I} = \{i^{u}_1, \ldots, i^{u}_{K^{u}}\}$ is finite, we set $c=\max_{i \in \mathcal{I}} c_i$. Then we have $\forall X \in \mathcal{X}$, $\forall f \in \mathcal{F}$, there exist $n$-dimension $W \in \mathcal{W}$ corresponding to $X$, such that $\max_{i \in \mathcal{I}} \mathbb{P}_{W}\left(R_{W \mid f(X)=i}\right)<c^n$.
  
  In $m$-dimension subspace $\mathcal{Z}\subset \mathbb{R}^m$, given a $K$-$\epsilon$-separable r.v. $Z \in \mathcal{Z}$ with $\mathcal{F}=\{f: \mathcal{X} \rightarrow \mathcal{I}\}$, then $X \in \mathcal{X}$, $\forall f \in \mathcal{F}$, $\tau(Z, f(X))=\max_{i, j \in \mathcal{I}, i \neq j} \mathbb{P}_{Z}\left(R_{Z \mid f(X)=i} \cap R_{Z \mid f(X)=j}\right) = \epsilon$. Based on the $m$-dimension r.v. $Z \in \mathcal{Z}$ corresponding to $X$, the $n$-dimension expansion $W \in \mathcal{W}$ must satisfy that 
  $\max_{i, j \in \mathcal{I}, i \neq j} \mathbb{P}_{W}\left(R_{W \mid f(X)=i} \cap R_{W \mid f(X)=j} \right)<\epsilon c^{n-m}$.
  Thus, $W$ is $K$-$\epsilon$-separable with the same $\mathcal{F}=\{f: \mathcal{X} \rightarrow \mathcal{I}\}$.\\

  \textbf{Not Necessity:} It is easy to prove if we find a specific counterexample to show that based on a $K$-$\epsilon$-separable $n$-dimension r.v. $W \in \mathcal{W} \subset \mathcal{X}$, a $m$-dimension projection r.v. $Z \in \mathcal{Z} \subset \mathcal{W}$ is not $K$-$\epsilon$-separable.

  Given a $3$-dimension space $\mathcal{X}=\{(x,y,z),x\in[-1,1], y\in[-1,1], z\in[-1,1]\}\subset\mathbb{R}^3$ and r.v. $X$ is randomly sampled from $\mathcal{X}$ with uniform distribution. We assume that the label of $X$ is 1 if $X$ is located on or above the $2$-dimension surface $z = x^2$ and if $\mathcal{X}$ is below the surface the label is 0. Then we define a classification function $f(X):=0$ if $z <x^2$ and $f(X):=1$ if $z \geq x^2$. As $\mathbb{P}_{X}\left(R_{X \mid f(X)=0} \cap R_{X \mid f(X)=1}\right) = 0$, we have r.v. $X$ is $2$-$0$-separable with non-empty $\mathcal{F}=\{f: \mathcal{X} \rightarrow \mathcal{I}\}$, where $\mathcal{I} = \{0,1\}$.
  
  It could be easy to prove that in the $2$-dimension subspace $\mathcal{W}=\{(x,0,z),x\in[-1,1], z\in[-1,1]\}\subset\mathcal{X}$, r.v. $W$ is also $2$-$0$-separable with the same $\mathcal{F}=\{f: \mathcal{X} \rightarrow \mathcal{I}\}$, because $\mathbb{P}_{W}\left(R_{W \mid f(X)=0} \cap R_{W \mid f(X)=1}\right) = 0$ is still hold true. Then, we consider a $1$-dimension subspace $\mathcal{Z}=\{(x,0,0),x \in [-1,1]\} \subset \mathcal{W}$ and r.v. $Z\in \mathcal{Z}$ is a projection of $X$ on the $x$-axis. It is obvious that $\tau(Z, f(X))=\mathbb{P}_{Z}\left(R_{Z \mid f(X)=0} \cap R_{Z \mid f(X)=1}\right) = 2$ is a consistent value, so r.v. $Z$ is not $2$-$\epsilon$-separable in space $\mathcal{Z}$ with any $\epsilon < 2$.
\end{proof}

\subsection{Proof of Theorem 2}\label{proof_Theorem_2}
Here we will prove that the condition $\Pi^l_c \bigcap \Pi^u_c \neq \emptyset$ could not guarantee $\Pi^l \bigcap \Pi^u \neq \emptyset$.
\begin{proof}
  The transformations $\pi^l_s(X)$ and $\pi^u_s(X)$ encode exclusive style information of two domains respectively. As the distribution of datasets $X^l$ and $X^u$ from two domains are different, thus $\Pi^l_s \bigcap \Pi^u_s := \{\pi^l_s, \pi^l_s \in \Pi^l_s \} \bigcap \{\pi^u_s, \pi^u_s \in \Pi^u_s \} = \emptyset$. Even $\Pi^l_c \bigcap \Pi^u_c \neq \emptyset$, there still has $\Pi^l \bigcap \Pi^u = [\Pi^l_c, \Pi^l_s] \bigcap [\Pi^u_c, \Pi^u_s] := \{[\pi^l_c, \pi^l_s], \pi^l_c \in \Pi^l_c  \ {\rm and} \  \pi^l_s \in \Pi^l_s \} \bigcap \{[\pi^u_c, \pi^u_s], \pi^u_c \in \Pi^u_c \ {\rm and} \ \pi^u_s \in \Pi^u_s \} = \emptyset$.
\end{proof}

\subsection{NMI and ARI results on OfficeHome and DomainNet40}\label{NMI_ARI_test_results}

The results of cross domain NCD tasks evaluated by NMI and ARI on OfficeHome \cite{OfficeHome} and DomainNet40 \cite{DomainNet40} are shown in Tab. \ref{cross_domain_Officehome_NMI}, \ref{cross_domain_DomainNet40_NMI}, \ref{cross_domain_Officehome_ARI}, and \ref{cross_domain_DomainNet40_ARI}, respectively.

\begin{table*}[htbp]
  \begin{center}
  \caption{NMI with std(\%) on OfficeHome \cite{OfficeHome}. The best and second-best results are in bold and underlined, respectively. The upper arrow indicates there exists improvement with the assistance of proposed module compared with the corresponding original methods.}
  \label{cross_domain_Officehome_NMI}
  \resizebox{\linewidth}{!}{
  \begin{tabular}{l|lll|lll|lll|lll|l}
    \toprule
    ~ & $\textbf{R}\rightarrow\textbf{A}$ & $\textbf{R}\rightarrow\textbf{C}$ & $\textbf{R}\rightarrow\textbf{P}$ & $\textbf{A}\rightarrow\textbf{R}$ & $\textbf{A}\rightarrow\textbf{C}$ & $\textbf{A}\rightarrow\textbf{P}$ & $\textbf{C}\rightarrow\textbf{R}$ & $\textbf{C}\rightarrow\textbf{A}$ & $\textbf{C}\rightarrow\textbf{P}$ & $\textbf{P}\rightarrow\textbf{R}$ & $\textbf{P}\rightarrow\textbf{A}$ & $\textbf{P}\rightarrow\textbf{C}$ & \textbf{Mean} \\
    \midrule
    SimGCD\cite{SimGCD} & $0.7327_{(2.0)}$ & $\underline{0.5657}_{(2.9)}$ & $0.8005_{(2.1)}$ & $0.7928_{(1.1)}$ & $0.5731_{(2.4)}$ & $0.7882_{(2.6)}$ & $0.8115_{(1.3)}$ & $0.7157_{(0.7)}$ & $0.8033_{(2.0)}$ & $0.7932_{(1.2)}$ & $0.7285_{(1.8)}$ & $0.5574_{(1.7)}$ & $0.7219_{(1.7)}$
    \\
    \midrule
    $+\mathcal{L}_{\mathrm{orth}}$ & $0.7357_{(1.7)}$\textuparrow & $\mathbf{0.5696}_{(2.6)}$\textuparrow & $\mathbf{0.8104}_{(1.1)}$\textuparrow & $0.7945_{(1.2)}$\textuparrow & $0.5757_{(2.6)}$\textuparrow & $0.7887_{(1.6)}$\textuparrow & $0.8109_{(1.4)}$ & $\mathbf{0.7349}_{(1.9)}$\textuparrow & $\mathbf{0.8109}_{(1.1)}$\textuparrow & $\mathbf{0.7975}_{(1.8)}$\textuparrow & $\mathbf{0.7305}_{(2.0)}$\textuparrow & $0.5597_{(0.6)}$\textuparrow & $\mathbf{0.7266}_{(1.6)}$\textuparrow
    \\
    $+\mathcal{L}_{\mathrm{cossimi}}$ & $\underline{0.7377}_{(1.7)}$\textuparrow & $0.5656_{(3.1)}$ & $\underline{0.8036}_{(2.0)}$\textuparrow & $0.7948_{(1.2)}$\textuparrow & $0.5673_{(3.8)}$ & $\mathbf{0.7898}_{(2.3)}$\textuparrow & $\underline{0.8133}_{(1.7)}$\textuparrow & $\underline{0.7230}_{(0.7)}$\textuparrow & $\underline{0.8097}_{(1.8)}$\textuparrow & $0.7963_{(1.7)}$\textuparrow & $\underline{0.7303}_{(1.4)}$\textuparrow & $0.5631_{(1.9)}$\textuparrow & $\underline{0.7245}_{(1.9)}$\textuparrow
    \\
    $+\mathcal{L}_{\mathrm{corr}}$ & $\mathbf{0.7378}_{(1.8)}$\textuparrow & $0.5639_{(3.1)}$ & $0.8012_{(2.0)}$\textuparrow & $0.7954_{(1.3)}$\textuparrow & $0.5712_{(3.2)}$ & $\underline{0.7891}_{(1.9)}$\textuparrow & $\mathbf{0.8146}_{(1.5)}$\textuparrow & $0.7194_{(0.8)}$\textuparrow & $0.8035_{(2.1)}$\textuparrow & $\underline{0.7969}_{(1.6)}$\textuparrow & $0.7244_{(1.5)}$ & $0.5610_{(1.8)}$\textuparrow & $0.7232_{(1.9)}$\textuparrow\\
    \midrule
    \midrule
    UNO\cite{UNO} & $0.6826_{(1.2)}$ & $0.5638_{(2.1)}$ & $0.7827_{(1.3)}$ & $0.8018_{(0.5)}$ & $0.5750_{(0.7)}$ & $0.7877_{(1.0)}$ & $0.7883_{(1.0)}$ & $0.6785_{(1.4)}$ & $0.7745_{(0.8)}$ & $0.7940_{(1.2)}$ & $0.6893_{(0.7)}$ & $0.5682_{(0.5)}$ & $0.7072_{(1.0)}$ \\
    \midrule
    $+\mathcal{L}_{\mathrm{orth}}$ & $0.6826_{(0.4)}$ & $0.5574_{(2.2)}$ & $0.7799_{(1.1)}$ & $\mathbf{0.8043}_{(1.3)}$\textuparrow & $\mathbf{0.5847}_{(2.1)}$\textuparrow & $0.7855_{(1.2)}$ & $0.7774_{(1.2)}$ & $0.6836_{(0.7)}$\textuparrow & $0.7747_{(0.3)}$\textuparrow & $0.7962_{(1.1)}$\textuparrow & $0.6839_{(1.1)}$ & $0.5652_{(0.7)}$ & $0.7063_{(1.1)}$ \\
    $+\mathcal{L}_{\mathrm{cossimi}}$ & $0.6857_{(1.4)}$\textuparrow & $0.5601_{(3.0)}$ & $0.7822_{(1.5)}$ & $0.8022_{(0.9)}$\textuparrow & $\underline{0.5795}_{(1.3)}$\textuparrow & $0.7868_{(0.7)}$ & $0.7889_{(0.7)}$\textuparrow & $0.6842_{(0.8)}$\textuparrow & $0.7744_{(0.7)}$ & $0.7954_{(1.3)}$\textuparrow & $0.6846_{(0.8)}$ & $0.5667_{(1.4)}$ & $0.7075_{(1.2)}$\textuparrow \\
    $+\mathcal{L}_{\mathrm{corr}}$ & $0.6848_{(1.1)}$\textuparrow & $0.5618_{(2.7)}$ & $0.7813_{(1.5)}$ & $\underline{0.8030}_{(0.8)}$\textuparrow & $0.5789_{(1.2)}$\textuparrow & $0.7872_{(0.7)}$ & $0.7896_{(1.0)}$\textuparrow & $0.6840_{(0.8)}$\textuparrow & $0.7744_{(0.7)}$ & $0.7935_{(1.3)}$ & $0.6852_{(0.9)}$ & $0.5667_{(1.3)}$ & $0.7075_{(1.2)}$\textuparrow \\
    \midrule
    \midrule
    ComEx\cite{ComEx} & $0.6287_{(0.6)}$ & $0.5559_{(1.2)}$ & $0.7629_{(1.1)}$ & $0.7872_{(1.1)}$ & $0.5737_{(1.7)}$ & $0.7742_{(1.2)}$ & $0.7815_{(2.4)}$ & $0.6365_{(1.4)}$ & $0.7663_{(1.1)}$ & $0.7948_{(1.0)}$ & $0.6207_{(3.3)}$ & $0.5692_{(1.5)}$ & $0.6876_{(1.5)}$\\
    \midrule
    $+\mathcal{L}_{\mathrm{orth}}$ & $0.6260_{(0.3)}$ & $0.5568_{(1.6)}$\textuparrow & $0.7603_{(2.1)}$ & $0.7914_{(1.7)}$\textuparrow & $0.5759_{(2.0)}$\textuparrow & $0.7780_{(1.5)}$\textuparrow & $0.7809_{(2.1)}$ & $0.6424_{(0.8)}$\textuparrow & $0.7734_{(1.5)}$\textuparrow & $0.7882_{(1.5)}$ & $0.6263_{(2.3)}$\textuparrow & $0.5611_{(1.3)}$ & $0.6884_{(1.6)}$\textuparrow \\
    $+\mathcal{L}_{\mathrm{cossimi}}$ & $0.6267_{(0.6)}$ & $0.5636_{(1.6)}$\textuparrow & $0.7533_{(2.1)}$ & $0.7886_{(1.3)}$\textuparrow & $0.5781_{(0.9)}$\textuparrow & $0.7780_{(0.6)}$\textuparrow & $0.7778_{(1.8)}$ & $0.6336_{(0.7)}$ & $0.7663_{(1.2)}$ & $0.7896_{(1.5)}$ & $0.6253_{(2.2)}$\textuparrow & $\mathbf{0.5815}_{(1.2)}$\textuparrow & $0.6885_{(1.3)}$\textuparrow \\
    $+\mathcal{L}_{\mathrm{corr}}$ & $0.6304_{(0.8)}$\textuparrow & $0.5561_{(1.5)}$\textuparrow & $0.7558_{(1.2)}$ & $0.7922_{(1.6)}$\textuparrow & $0.5778_{(1.0)}$\textuparrow & $0.7788_{(1.0)}$\textuparrow & $0.7815_{(2.3)}$ & $0.6338_{(0.5)}$ & $0.7635_{(0.9)}$ & $0.7851_{(1.6)}$ & $0.6218_{(2.3)}$\textuparrow & $\underline{0.5810}_{(0.7)}$\textuparrow & $0.6882_{(1.3)}$\textuparrow \\
    \bottomrule
  \end{tabular}}
  \end{center}
\end{table*}

\begin{table*}[htbp]
  \begin{center}
  \caption{ARI with std(\%) on OfficeHome \cite{OfficeHome}.}
  \label{cross_domain_Officehome_ARI}
  \resizebox{\linewidth}{!}{
  \begin{tabular}{l|lll|lll|lll|lll|l}
    \toprule
    ~ & $\textbf{R}\rightarrow\textbf{A}$ & $\textbf{R}\rightarrow\textbf{C}$ & $\textbf{R}\rightarrow\textbf{P}$ & $\textbf{A}\rightarrow\textbf{R}$ & $\textbf{A}\rightarrow\textbf{C}$ & $\textbf{A}\rightarrow\textbf{P}$ & $\textbf{C}\rightarrow\textbf{R}$ & $\textbf{C}\rightarrow\textbf{A}$ & $\textbf{C}\rightarrow\textbf{P}$ & $\textbf{P}\rightarrow\textbf{R}$ & $\textbf{P}\rightarrow\textbf{A}$ & $\textbf{P}\rightarrow\textbf{C}$ & \textbf{Mean} \\
    \midrule
    SimGCD\cite{SimGCD} & $0.4778_{(5.0)}$ & $0.2268_{(3.2)}$ & $0.6620_{(5.3)}$ & $0.6402_{(3.3)}$ & $0.2516_{(3.2)}$ & $0.6418_{(4.7)}$ & $0.6804_{(2.1)}$ & $0.4227_{(5.2)}$ & $0.6422_{(4.9)}$ & $0.6451_{(2.6)}$ & $\underline{0.4966}_{(3.9)}$ & $0.2061_{(2.3)}$ & $0.4994_{(3.8)}$\\
    \midrule
    $+\mathcal{L}_{\mathrm{orth}}$ & $0.4904_{(3.3)}$\textuparrow & $0.2281_{(2.7)}$\textuparrow & $\mathbf{0.6680}_{(2.2)}$\textuparrow & $\underline{0.6464}_{(2.8)}$\textuparrow & $0.2576_{(4.0)}$\textuparrow & $0.6427_{(4.6)}$\textuparrow & $\underline{0.6822}_{(2.5)}$\textuparrow & $\mathbf{0.4883}_{(3.4)}$\textuparrow & $\mathbf{0.6829}_{(2.5)}$\textuparrow & $\underline{0.6512}_{(3.1)}$\textuparrow & $0.4965_{(4.9)}$ & $0.2257_{(2.1)}$\textuparrow & $\mathbf{0.5133}_{(3.2)}$\textuparrow\\
    $+\mathcal{L}_{\mathrm{cossimi}}$ & $\mathbf{0.4931}_{(2.9)}$\textuparrow & $0.2232_{(3.8)}$ & $\underline{0.6649}_{(5.4)}$\textuparrow & $0.6434_{(3.6)}$\textuparrow & $0.2525_{(4.4)}$\textuparrow & $\underline{0.6447}_{(5.4)}$\textuparrow & $0.6817_{(2.5)}$\textuparrow & $\underline{0.4425}_{(4.5)}$\textuparrow & $\underline{0.6585}_{(4.2)}$\textuparrow & $0.6489_{(2.7)}$\textuparrow & $\mathbf{0.4997}_{(3.8)}$\textuparrow & $0.2160_{(2.2)}$\textuparrow & $\underline{0.5058}_{(3.8)}$\textuparrow
    \\
    $+\mathcal{L}_{\mathrm{corr}}$ & $\underline{0.4923}_{(2.8)}$\textuparrow & $0.2228_{(3.6)}$ & $0.6619_{(5.3)}$ & $0.6442_{(3.7)}$\textuparrow & $0.2520_{(4.2)}$\textuparrow & $\mathbf{0.6455}_{(5.1)}$\textuparrow & $\mathbf{0.6845}_{(2.4)}$\textuparrow & $0.4424_{(4.5)}$\textuparrow & $0.6463_{(5.4)}$\textuparrow & $0.6511_{(3.1)}$\textuparrow & $0.4933_{(2.5)}$ & $0.2160_{(2.5)}$\textuparrow & $0.5044_{(3.8)}$\textuparrow\\
    \midrule
    \midrule
    UNO\cite{UNO} & $0.3674_{(1.5)}$ & $0.2503_{(2.3)}$ & $0.5765_{(3.2)}$ & $0.6472_{(1.1)}$ & $0.2693_{(0.9)}$ & $0.6025_{(1.4)}$ & $0.6184_{(2.4)}$ & $0.3641_{(3.1)}$ & $0.5568_{(1.8)}$ & $0.6313_{(1.5)}$ & $0.3835_{(2.1)}$ & $0.2586_{(1.0)}$ & $0.4605_{(1.9)}$ \\
    \midrule
    $+\mathcal{L}_{\mathrm{orth}}$ & $0.3703_{(1.6)}$\textuparrow & $0.2462_{(2.7)}$ & $0.5732_{(2.5)}$ & $0.6490_{(2.5)}$\textuparrow & $0.2802_{(2.4)}$\textuparrow & $0.5969_{(3.0)}$ & $0.6001_{(2.2)}$ & $0.3744_{(1.4)}$\textuparrow & $0.5586_{(0.6)}$\textuparrow & $0.6372_{(1.5)}$\textuparrow & $0.3682_{(1.5)}$ & $0.2560_{(1.5)}$ & $0.4592_{(1.9)}$ \\
    $+\mathcal{L}_{\mathrm{cossimi}}$ & $0.3770_{(2.4)}$\textuparrow & $0.2460_{(3.2)}$ & $0.5761_{(3.7)}$ & $0.6450_{(1.7)}$ & $0.2719_{(1.0)}$\textuparrow & $0.5999_{(0.9)}$ & $0.6191_{(2.2)}$\textuparrow & $0.3733_{(1.8)}$\textuparrow & $0.5566_{(1.8)}$ & $0.6351_{(1.8)}$\textuparrow & $0.3755_{(2.0)}$ & $0.2599_{(1.4)}$\textuparrow & $0.4613_{(2.0)}$\textuparrow \\
    $+\mathcal{L}_{\mathrm{corr}}$ & $0.3734_{(1.9)}$\textuparrow & $0.2471_{(3.0)}$ & $0.5764_{(3.7)}$ & $0.6471_{(1.4)}$ & $0.2708_{(1.0)}$\textuparrow & $0.6004_{(1.0)}$ & $0.6203_{(2.7)}$\textuparrow & $0.3726_{(1.9)}$\textuparrow & $0.5565_{(1.8)}$ & $0.6313_{(1.9)}$\textuparrow & $0.3775_{(2.0)}$ & $0.2603_{(1.3)}$\textuparrow & $0.4611_{(2.0)}$\textuparrow \\
    \midrule
    \midrule
    ComEx\cite{ComEx} & $0.3207_{(1.0)}$ & $0.2524_{(2.1)}$ & $0.5632_{(1.8)}$ & $0.6407_{(1.6)}$ & $0.2895_{(2.5)}$ & $0.6015_{(2.6)}$ & $0.6334_{(4.8)}$ & $0.3206_{(2.6)}$ & $0.5842_{(2.6)}$ & $\mathbf{0.6582}_{(2.0)}$ & $0.3008_{(5.3)}$ & $0.2772_{(2.2)}$ & $0.4535_{(2.6)}$ \\
    \midrule
    $+\mathcal{L}_{\mathrm{orth}}$ & $0.3160_{(1.7)}$ & $0.2546_{(1.9)}$\textuparrow & $0.5588_{(4.0)}$ & $\mathbf{0.6554}_{(2.7)}$\textuparrow & $0.2872_{(2.3)}$ & $0.5963_{(3.4)}$ & $0.6338_{(4.3)}$\textuparrow & $0.3345_{(1.6)}$\textuparrow & $0.5887_{(3.8)}$\textuparrow & $0.6447_{(2.1)}$ & $0.3075_{(3.8)}$\textuparrow & $0.2648_{(1.6)}$ & $0.4535_{(2.7)}$\textuparrow \\
    $+\mathcal{L}_{\mathrm{cossimi}}$ & $0.3185_{(0.5)}$ & $\mathbf{0.2605}_{(2.2)}$\textuparrow & $0.5496_{(3.3)}$ & $0.6431_{(2.3)}$\textuparrow & $\underline{0.2927}_{(1.8)}$\textuparrow & $0.6091_{(1.0)}$\textuparrow & $0.6331_{(3.8)}$ & $0.3188_{(1.4)}$ & $0.5839_{(3.0)}$ & $0.6470_{(2.8)}$ & $0.3045_{(4.1)}$\textuparrow & $\underline{0.2930}_{(2.4)}$\textuparrow & $0.4545_{(2.4)}$\textuparrow \\
    $+\mathcal{L}_{\mathrm{corr}}$ & $0.3235_{(0.6)}$\textuparrow & $\underline{0.2549}_{(1.9)}$\textuparrow & $0.5544_{(1.7)}$ & $\underline{0.6516}_{(2.2)}$\textuparrow & $\mathbf{0.2948}_{(1.9)}$\textuparrow & $0.6095_{(1.6)}$\textuparrow & $0.6380_{(4.8)}$\textuparrow & $0.3164_{(0.8)}$ & $0.5790_{(2.4)}$ & $0.6367_{(2.6)}$ & $0.2981_{(4.4)}$ & $\mathbf{0.2932}_{(1.6)}$\textuparrow & $0.4542_{(2.2)}$\textuparrow \\
    \bottomrule
  \end{tabular}}
  \end{center}
\end{table*}

\begin{table*}[htbp]
  \begin{center}
  \caption{NMI with std(\%) on DomainNet40 \cite{DomainNet40}.}
  \label{cross_domain_DomainNet40_NMI}
  \resizebox{\linewidth}{!}{
  \begin{tabular}{l|lll|lll|lll|lll|l}
    \toprule
    ~ & $\textbf{R}\rightarrow\textbf{C}$ & $\textbf{R}\rightarrow\textbf{P}$ & $\textbf{R}\rightarrow\textbf{S}$ & $\textbf{C}\rightarrow\textbf{R}$ & $\textbf{C}\rightarrow\textbf{P}$ & $\textbf{C}\rightarrow\textbf{S}$ & $\textbf{P}\rightarrow\textbf{R}$ & $\textbf{P}\rightarrow\textbf{C}$ & $\textbf{P}\rightarrow\textbf{S}$ & $\textbf{S}\rightarrow\textbf{R}$ & $\textbf{S}\rightarrow\textbf{C}$ & $\textbf{S}\rightarrow\textbf{P}$ & \textbf{Mean} \\
    \midrule
    SimGCD\cite{SimGCD} & $\underline{0.5936}_{(1.3)}$ & $\mathbf{0.6706}_{(0.6)}$ & $\underline{0.4719}_{(1.6)}$ & $0.8497_{(0.2)}$ & $0.6529_{(1.8)}$ & $0.4678_{(1.5)}$ & $0.8460_{(1.1)}$ & $0.5843_{(1.7)}$ & $0.4709_{(0.7)}$ & $0.8534_{(0.4)}$ & $0.6029_{(2.8)}$ & $\mathbf{0.6747}_{(1.2)}$ & $\underline{0.6449}_{(1.2)}$
    \\
    \midrule
    $+\mathcal{L}_{\mathrm{orth}}$ & $\mathbf{0.6012}_{(0.6)}$\textuparrow & $0.6624_{(1.0)}$ & $\mathbf{0.4733}_{(0.9)}$\textuparrow & $0.8528_{(0.5)}$\textuparrow & $0.6592_{(1.8)}$\textuparrow & $0.4739_{(1.0)}$\textuparrow & $0.8484_{(1.6)}$\textuparrow & $0.5870_{(1.8)}$\textuparrow & $0.4710_{(1.1)}$\textuparrow & $0.8537_{(0.3)}$\textuparrow & $0.5996_{(2.7)}$ & $\underline{0.6743}_{(1.3)}$ & $\mathbf{0.6464}_{(1.2)}$\textuparrow\\
    $+\mathcal{L}_{\mathrm{cossimi}}$ & $0.5887_{(1.7)}$ & $\underline{0.6641}_{(0.8)}$ & $0.4685_{(1.6)}$ & $0.8491_{(0.3)}$ & $0.6537_{(1.8)}$\textuparrow & $0.4688_{(1.5)}$\textuparrow & $0.8452_{(1.1)}$ & $0.5846_{(1.8)}$\textuparrow & $0.4686_{(0.7)}$ & $0.8545_{(0.4)}$\textuparrow & $0.6026_{(2.3)}$ & $0.6739_{(1.2)}$ & $0.6435_{(1.3)}$\\
    $+\mathcal{L}_{\mathrm{corr}}$ & $0.5863_{(1.6)}$ & $0.6634_{(0.9)}$ & $0.4707_{(1.7)}$ & $0.8499_{(0.3)}$\textuparrow & $0.6537_{(2.0)}$\textuparrow & $0.4686_{(1.2)}$\textuparrow & $0.8455_{(1.1)}$ & $0.5834_{(1.8)}$ & $0.4699_{(0.7)}$ & $0.8544_{(0.5)}$\textuparrow & $0.5998_{(2.3)}$ & $0.6740_{(1.3)}$ & $0.6433_{(1.3)}$\\
    \midrule
    \midrule
    UNO\cite{UNO} & $0.5701_{(1.7)}$ & $0.6138_{(1.1)}$ & $0.4510_{(1.2)}$ & $0.8590_{(1.0)}$ & $0.6460_{(0.6)}$ & $0.4892_{(0.5)}$ & $0.8507_{(0.8)}$ & $\mathbf{0.5931}_{(1.4)}$ & $0.4791_{(0.3)}$ & $0.8481_{(1.0)}$ & $\underline{0.6189}_{(0.5)}$ & $0.6430_{(1.6)}$ & $0.6385_{(1.0)}$ \\
    \midrule
    $+\mathcal{L}_{\mathrm{orth}}$ & $0.5663_{(1.5)}$ & $0.6143_{(1.3)}$\textuparrow & $0.4517_{(1.4)}$\textuparrow & $0.8557_{(1.2)}$ & $0.6446_{(1.1)}$ & $0.4859_{(1.2)}$ & $0.8466_{(0.7)}$ & $0.5893_{(1.7)}$ & $0.4761_{(1.1)}$ & $0.8462_{(0.8)}$ & $0.6125_{(0.7)}$ & $0.6391_{(1.2)}$ & $0.6357_{(1.2)}$ \\
    $+\mathcal{L}_{\mathrm{cossimi}}$ & $0.5719_{(1.5)}$\textuparrow & $0.6119_{(0.7)}$ & $0.4567_{(1.4)}$\textuparrow & $0.8589_{(1.0)}$ & $0.6436_{(0.6)}$ & $\underline{0.4895}_{(0.7)}$\textuparrow & $0.8512_{(0.7)}$\textuparrow & $\underline{0.5922}_{(1.5)}$ & $\mathbf{0.4805}_{(0.6)}$\textuparrow & $0.8509_{(1.2)}$\textuparrow & $0.6187_{(0.6)}$ & $0.6410_{(1.4)}$ & $0.6389_{(1.0)}$\textuparrow \\
    $+\mathcal{L}_{\mathrm{corr}}$ & $0.5732_{(1.5)}$\textuparrow & $0.6128_{(1.0)}$ & $0.4572_{(1.4)}$\textuparrow& $0.8586_{(1.0)}$ & $0.6438_{(0.6)}$ & $\mathbf{0.4907}_{(0.6)}$\textuparrow & $0.8513_{(0.7)}$\textuparrow & $0.5917_{(1.5)}$ & $\underline{0.4793}_{(0.5)}$\textuparrow & $0.8505_{(1.3)}$\textuparrow & $\mathbf{0.6202}_{(0.4)}$\textuparrow & $0.6408_{(1.5)}$ & $0.6392_{(1.0)}$\textuparrow \\
    \midrule
    \midrule
    ComEx\cite{ComEx} & $0.5669_{(1.3)}$ & $0.6500_{(0.3)}$ & $0.4574_{(1.3)}$ & $0.8652_{(1.3)}$ & $\mathbf{0.6607}_{(0.7)}$ & $0.4784_{(1.3)}$ & $\underline{0.8664}_{(1.5)}$ & $0.5662_{(1.4)}$ & $0.4716_{(1.1)}$ & $0.8612_{(1.6)}$ & $0.5642_{(1.1)}$ & $0.6603_{(0.9)}$ & $0.6390_{(1.1)}$  \\
    \midrule
    $+\mathcal{L}_{\mathrm{orth}}$ & $0.5585_{(0.5)}$ & $0.6453_{(1.5)}$ & $0.4629_{(1.4)}$\textuparrow & $\underline{0.8663}_{(1.4)}$\textuparrow& $0.6593_{(0.5)}$ & $0.4796_{(1.0)}$\textuparrow & $\mathbf{0.8667}_{(1.1)}$\textuparrow & $0.5656_{(1.3)}$ & $0.4699_{(1.7)}$ & $0.8624_{(1.7)}$\textuparrow & $0.5693_{(1.5)}$\textuparrow & $0.6655_{(1.2)}$\textuparrow & $0.6393_{(1.2)}$\textuparrow \\
    $+\mathcal{L}_{\mathrm{cossimi}}$ & $0.5590_{(0.7)}$ & $0.6535_{(0.6)}$\textuparrow & $0.4631_{(2.1)}$\textuparrow & $\mathbf{0.8667}_{(1.4)}$\textuparrow & $\underline{0.6601}_{(0.7)}$ & $0.4826_{(1.0)}$\textuparrow & $0.8614_{(1.4)}$ & $0.5705_{(1.1)}$\textuparrow & $0.4724_{(1.6)}$\textuparrow & $\mathbf{0.8696}_{(0.6)}$\textuparrow & $0.5637_{(0.8)}$ & $0.6565_{(0.5)}$ & $0.6399_{(1.0)}$\textuparrow \\
    $+\mathcal{L}_{\mathrm{corr}}$ & $0.5662_{(1.6)}$ & $0.6529_{(0.3)}$\textuparrow & $0.4645_{(2.2)}$\textuparrow & $0.8660_{(1.2)}$\textuparrow & $0.6582_{(0.5)}$ & $0.4816_{(0.9)}$\textuparrow & $0.8618_{(1.4)}$ & $0.5730_{(0.8)}$\textuparrow & $0.4708_{(1.1)}$ & $\underline{0.8667}_{(0.7)}$\textuparrow & $0.5655_{(0.8)}$\textuparrow & $0.6601_{(0.8)}$ & $0.6406_{(1.0)}$\textuparrow \\
    \bottomrule
  \end{tabular}}
  \end{center}
\end{table*}

\begin{table*}[htbp]
  \begin{center}
  \caption{ARI with std(\%) on DomainNet40 \cite{DomainNet40}.}
  \label{cross_domain_DomainNet40_ARI}
  \resizebox{\linewidth}{!}{
  \begin{tabular}{l|lll|lll|lll|lll|l}
    \toprule
    ~ & $\textbf{R}\rightarrow\textbf{C}$ & $\textbf{R}\rightarrow\textbf{P}$ & $\textbf{R}\rightarrow\textbf{S}$ & $\textbf{C}\rightarrow\textbf{R}$ & $\textbf{C}\rightarrow\textbf{P}$ & $\textbf{C}\rightarrow\textbf{S}$ & $\textbf{P}\rightarrow\textbf{R}$ & $\textbf{P}\rightarrow\textbf{C}$ & $\textbf{P}\rightarrow\textbf{S}$ & $\textbf{S}\rightarrow\textbf{R}$ & $\textbf{S}\rightarrow\textbf{C}$ & $\textbf{S}\rightarrow\textbf{P}$ & \textbf{Mean} \\
    \midrule
    SimGCD \cite{SimGCD} & $\underline{0.3757}_{(1.9)}$ & $0.5239_{(3.9)}$ & $0.2340_{(1.2)}$ & $0.7859_{(0.7)}$ & $0.5580_{(4.2)}$ & $0.2316_{(2.1)}$ & $0.7751_{(2.5)}$ & $0.3748_{(3.4)}$ & $0.2367_{(0.3)}$ & $0.8011_{(1.6)}$ & $0.3827_{(4.2)}$ & $0.5660_{(4.0)}$ & $0.4871_{(2.5)}$\\
    \midrule
    $+\mathcal{L}_{\mathrm{orth}}$ & $\mathbf{0.3836}_{(1.4)}$\textuparrow & $0.5315_{(4.9)}$\textuparrow & $0.2340_{(1.0)}$ & $0.7943_{(1.8)}$\textuparrow & $\mathbf{0.5639}_{(3.5)}$\textuparrow & $0.2406_{(1.6)}$\textuparrow & $0.7889_{(2.9)}$\textuparrow & $0.3851_{(4.2)}$\textuparrow & $0.2382_{(0.4)}$\textuparrow & $0.8014_{(1.4)}$\textuparrow & $0.3816_{(4.5)}$ & $\mathbf{0.5724}_{(3.7)}$\textuparrow & $\mathbf{0.4930}_{(2.6)}$\textuparrow
    \\
    $+\mathcal{L}_{\mathrm{cossimi}}$ & $0.3656_{(2.2)}$ & $\underline{0.5319}_{(5.1)}$\textuparrow & $0.2296_{(1.7)}$ & $0.7842_{(0.8)}$ & $\underline{0.5597}_{(4.0)}$\textuparrow & $0.2336_{(2.1)}$\textuparrow & $0.7755_{(2.4)}$\textuparrow & $0.3759_{(3.5)}$\textuparrow & $0.2375_{(0.3)}$\textuparrow & $0.8025_{(1.5)}$\textuparrow & $0.3811_{(4.0)}$ & $0.5670_{(4.1)}$\textuparrow & $0.4870_{(2.6)}$
    \\
    $+\mathcal{L}_{\mathrm{corr}}$ & $0.3656_{(2.1)}$ & $\mathbf{0.5319}_{(5.0)}$\textuparrow & $0.2296_{(1.8)}$ & $0.7842_{(0.8)}$ & $0.5586_{(4.2)}$\textuparrow & $0.2331_{(1.8)}$\textuparrow & $0.7738_{(2.5)}$ & $0.3733_{(3.4)}$ & $0.2375_{(0.3)}$\textuparrow & $0.8011_{(1.5)}$ & $0.3811_{(4.0)}$ & $\underline{0.5670}_{(3.9)}$\textuparrow & $0.4864_{(2.6)}$\\
    \midrule
    \midrule
    UNO\cite{UNO} & $0.3649_{(2.2)}$ & $0.4269_{(1.4)}$ & $0.2468_{(1.3)}$ & $0.7908_{(1.8)}$ & $0.4490_{(1.6)}$ & $0.2734_{(0.7)}$ & $0.7750_{(1.6)}$ & $\underline{0.3887}_{(1.9)}$ & $0.2624_{(0.8)}$ & $0.7664_{(2.2)}$ & $0.4206_{(1.0)}$ & $0.4603_{(1.7)}$ & $0.4688_{(1.5)}$ \\
    \midrule
    $+\mathcal{L}_{\mathrm{orth}}$ & $0.3608_{(1.5)}$ & $0.4302_{(1.6)}$\textuparrow & $0.2448_{(1.2)}$ & $0.7843_{(2.1)}$ & $0.4510_{(1.6)}$\textuparrow & $0.2698_{(1.5)}$ & $0.7653_{(1.3)}$ & $0.3842_{(2.3)}$ & $0.2600_{(1.1)}$ & $0.7613_{(1.8)}$ & $0.4139_{(2.0)}$ & $0.4532_{(1.0)}$ & $0.4649_{(1.6)}$ \\
    $+\mathcal{L}_{\mathrm{cossimi}}$ & $0.3690_{(1.8)}$\textuparrow & $0.4253_{(0.9)}$ & $0.2519_{(1.4)}$\textuparrow & $0.7911_{(1.8)}$\textuparrow & $0.4461_{(1.4)}$ & $0.2740_{(0.8)}$\textuparrow & $0.7768_{(1.4)}$\textuparrow & $\mathbf{0.3896}_{(1.8)}$\textuparrow & $\mathbf{0.2649}_{(1.0)}$\textuparrow & $0.7723_{(2.5)}$\textuparrow & $\underline{0.4228}_{(1.0)}$\textuparrow & $0.4566_{(1.5)}$ & $0.4700_{(1.4)}$\textuparrow \\
    $+\mathcal{L}_{\mathrm{corr}}$ & $0.3716_{(1.6)}$\textuparrow & $0.4268_{(1.0)}$ & $0.2533_{(1.5)}$\textuparrow & $0.7907_{(1.9)}$ & $0.4456_{(1.4)}$ & $0.2747_{(0.8)}$\textuparrow & $0.7769_{(1.4)}$\textuparrow & $0.3882_{(1.9)}$ & $\underline{0.2631}_{(1.0)}$\textuparrow & $0.7721_{(2.6)}$\textuparrow & $\mathbf{0.4259}_{(1.1)}$\textuparrow & $0.4564_{(1.7)}$ & $0.4704_{(1.5)}$\textuparrow \\
    \midrule
    \midrule
    ComEx\cite{ComEx} & $0.3644_{(1.9)}$ & $0.5031_{(0.8)}$ & $0.2469_{(1.6)}$ & $0.8176_{(2.8)}$ & $0.5185_{(1.4)}$ & $0.2768_{(1.8)}$ & $\underline{0.8133}_{(2.8)}$ & $0.3653_{(2.2)}$ & $0.2604_{(1.7)}$ & $0.8042_{(3.4)}$ & $0.3713_{(1.6)}$ & $0.5163_{(1.2)}$ & $0.4882_{(1.9)}$ \\
    \midrule
    $+\mathcal{L}_{\mathrm{orth}}$ & $0.3513_{(1.3)}$ & $0.5008_{(2.9)}$ & $0.2523_{(2.0)}$\textuparrow & $\underline{0.8199}_{(3.5)}$\textuparrow & $0.5139_{(1.5)}$ & $0.2749_{(1.3)}$ & $\mathbf{0.8136}_{(1.8)}$\textuparrow & $0.3619_{(1.8)}$ & $0.2568_{(2.4)}$ & $0.8048_{(3.2)}$\textuparrow & $0.3726_{(2.7)}$\textuparrow & $0.5193_{(1.4)}$\textuparrow & $0.4868_{(2.1)}$ \\
    $+\mathcal{L}_{\mathrm{cossimi}}$ & $0.3553_{(1.4)}$ & $0.5082_{(1.0)}$\textuparrow & $\mathbf{0.2553}_{(2.6)}$\textuparrow & $\mathbf{0.8219}_{(3.1)}$\textuparrow & $0.5146_{(1.7)}$ & $\mathbf{0.2796}_{(1.4)}$\textuparrow & $0.8010_{(2.6)}$ & $0.3738_{(1.9)}$\textuparrow & $0.2583_{(1.9)}$ & $\mathbf{0.8209}_{(1.5)}$\textuparrow & $0.3705_{(1.3)}$ & $0.5117_{(0.6)}$ & $0.4893_{(1.7)}$\textuparrow \\
    $+\mathcal{L}_{\mathrm{corr}}$ & $0.3639_{(2.1)}$ & $0.5095_{(0.7)}$\textuparrow & $\underline{0.2536}_{(2.9)}$\textuparrow & $0.8181_{(2.6)}$\textuparrow & $0.5131_{(1.2)}$ & $\underline{0.2787}_{(1.2)}$\textuparrow & $0.8022_{(2.3)}$ & $0.3746_{(1.3)}$\textuparrow & $0.2595_{(1.2)}$ & $\underline{0.8133}_{(1.5)}$\textuparrow & $0.3693_{(1.1)}$ & $0.5180_{(1.4)}$\textuparrow & $\underline{0.4895}_{(1.6)}$\textuparrow \\
    \bottomrule
  \end{tabular}}
  \end{center}
\end{table*}

\end{CJK}
\end{document}